%% file: colm2026_conference.tex
\documentclass{article} % For LaTeX2e
\usepackage[final]{colm2026_conference}

\usepackage{microtype}
\usepackage{hyperref}
\usepackage{url}
\usepackage{booktabs}

\usepackage{hyperref}
\usepackage{url}

\usepackage{soul}
\usepackage{algorithm}
\usepackage{algpseudocode}
\usepackage{xcolor}
\usepackage{todonotes}
\usepackage{microtype}
\usepackage{hyperref}
\usepackage{url}
\usepackage{booktabs}
\usepackage{graphicx}
\usepackage{lineno}
\usepackage{enumitem}
\usepackage{multirow,colortbl,amssymb}
\usepackage{multirow}
\usepackage{threeparttable}
\usepackage{amsmath} 
\usepackage{wrapfig}
\usepackage{float}
\usepackage{caption}
\usepackage{subcaption}
\usepackage{afterpage} 
\usepackage{makecell}
\usepackage{array}
\usepackage{booktabs}
\usepackage[table]{xcolor}

\usepackage[nameinlink, capitalise]{cleveref}

\usepackage{amssymb}
\usepackage{pifont}
\usepackage{xcolor}

\definecolor{phaseblue}{RGB}{70,130,180}
\definecolor{phasegreen}{RGB}{60,140,90}
\definecolor{phaseorange}{RGB}{220,140,40}
\definecolor{phasered}{RGB}{200,80,80}

\definecolor{batchblue}{RGB}{40,100,160}
\definecolor{dbgreen}{RGB}{40,120,70}
\definecolor{qared}{RGB}{170,50,50}

\definecolor{revision}{HTML}{53579c}

\definecolor{bggray}{HTML}{959595}
\definecolor{lightgray}{rgb}{0.96,0.96,0.98} % Define a light gray color
\definecolor{lightpurple}{rgb}{0.91, 0.91, 0.96}

\definecolor{lightgray}{RGB}{245,245,245}
\definecolor{midgray}{RGB}{230,230,230}

\definecolor{bgpurple}{rgb}{0.95,0.95,0.98} % Define background purple color
\definecolor{darkgreen}{RGB}{0,100,0}
\definecolor{darkred}{rgb}{0.8, 0.25, 0.33}

\definecolor{darkgreen}{RGB}{0,100,0}
\definecolor{darkred}{rgb}{0.8, 0.25, 0.33}

\newcommand{\method}{\textsc{KBevo}}

\newcommand{\qwens}{\textsc{Qwen3-1.7B}}
\newcommand{\qwenm}{\textsc{Qwen3-4B}}

\usepackage{lineno}

\definecolor{darkblue}{rgb}{0, 0, 0.5}
\hypersetup{colorlinks=true, citecolor=darkblue, linkcolor=darkblue, urlcolor=darkblue}

\title{Co-Evolving Structured Knowledge and Reasoning \\in Language Models}

\author{Ryan Thomas Noonan$^{*}$ \And
Linxi Zhao$^{*,\dagger}$ \And
Menghan Xu$^{*}$ \And
Akanksha Sarkar \AND
Mihir Mishra \And
Dongyoung Go \And
Kilian Q. Weinberger \And
Yoav Artzi \And
Jennifer J. Sun \AND
\normalfont
Cornell University\\
\texttt{\{rtn27,lz586,mx253,as2637,mrm367,dg793,}\texttt{kilian,yoavartzi,jennifer.sun\}@cornell.edu}
}

\usepackage[most]{tcolorbox}
\usepackage{enumitem} 

\newtcolorbox{promptbox}{
  enhanced,
  colback=white,
  colframe=black!75,
  arc=2mm,
  boxrule=0.6pt,
  left=10pt, right=10pt,
  top=4pt, bottom=4pt,
  width=\linewidth,
  boxsep=0pt,
  breakable
}
\begin{document}

\begingroup
\renewcommand{\thefootnote}{}
\footnotetext{* Equal contribution. \quad $\dagger$ Project lead.}
\endgroup

\ifcolmsubmission
\linenumbers
\fi

% \maketitle
\begingroup
\setlength{\tabcolsep}{0pt}   % title block only; restored by \endgroup
\maketitle
\endgroup

\begin{abstract}
Retrieval-augmented methods improve factual accuracy by grounding language models in external knowledge, but retrieving over unstructured text often introduces irrelevant context and offers limited control over the retrieved information. Structured knowledge bases offer a more controllable alternative, yet they are expensive to construct and often brittle to reason over. To address these limitations, we propose \method{}: a \textit{co-evolving} framework that jointly learns to construct a structured knowledge base and reason over it for knowledge-intensive question answering. By optimizing both components end-to-end with QA outcome rewards, our method enables reasoning success to directly improve the quality of the constructed knowledge base.
This leads to larger, better-connected knowledge structures with higher answer reachability, while also improving compositional factual reasoning and controllability compared to standard retrieval baselines.
\begingroup
\renewcommand{\thefootnote}{\fnsymbol{footnote}}
\footnotetext[3]{We open-source our code and models at \url{https://github.com/kilian-group/KBevo}.} % 2 = dagger
\endgroup
\end{abstract}

\section{Introduction}

While language models have demonstrated strong reasoning and generation capabilities, their ability to reliably store and recall parametric factual knowledge remains limited. 
When knowledge is memorized during pretraining, the resulting representations are inherently lossy: 
facts may be partially memorized, conflated with related but distinct information~\citep{meng2022locating, bommasani2021opportunities}. Fundamentally, parametric knowledge is difficult to inspect, correct, or update without expensive retraining, making it ill-suited for knowledge-intensive applications where accuracy and controllability are essential.

Retrieval-augmented generation (RAG) and search agents have emerged as prominent approaches to address these limitations by grounding model outputs in external text corpora at inference time. By retrieving relevant passages and conditioning generation on them, these systems can access up-to-date information and reduce hallucination  \citep{lewis2020retrieval, izacard2021leveraging, li2025search, jin2025search, zheng2025deepresearcher}. 
The dominant paradigm, however, retrieves from unstructured text split into fixed-size chunks, a format optimized for coverage rather than reasoning \citep{karpukhin2020dense, jin2025flashrag}. As a result, retrieved chunks can inject irrelevant context or only partially address the query \citep{shi2023large, barnett2024seven, gao2025beyond, chang2026karl}. 

Structured knowledge bases (KBs) offer a promising alternative by representing facts as discrete, queryable entities and relations.
This representation enables precise access to individual facts and supports compositional reasoning, while making the stored knowledge more interpretable and editable \citep{vrandevcic2014wikidata, saxena2020improving, ho2020constructing, zhong2023mquake, kansal2026knowledge}. Because individual facts are stored explicitly, they can be modified without updating the model parameters. This gives users more direct control over the external knowledge available during reasoning.
However, KB coverage is often constrained by the schema specified at construction time, while building high-quality KBs requires substantial human annotation or costly frontier-model inference \citep{vrandevcic2014wikidata, dedhia2025bottom}. This construction cost makes structured KBs difficult to scale and build on demand.

\begin{figure}[!t]
\captionsetup{skip=5pt}
\centering
\includegraphics[width=\linewidth]{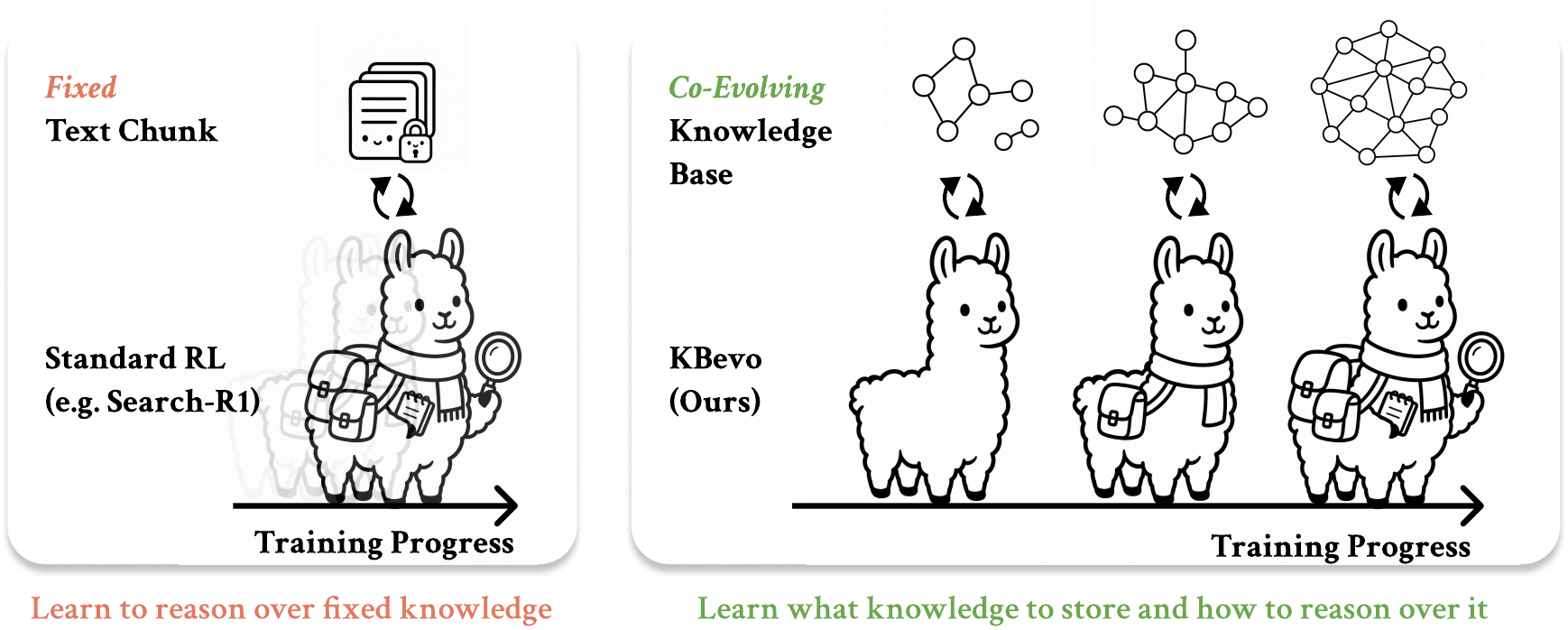}
\caption{\textbf{Co-evolving knowledge and reasoning.} Standard retrieval-based RL improves the reasoning policy while operating over fixed text chunks. In contrast, \method{} jointly optimizes reasoning and structured knowledge base construction, allowing the model and its structured knowledge base to improve together during training.
}
\label{fig:overview_top}
\vspace{-1em}
\end{figure}

The limitations of existing approaches share a common root: the decoupling of knowledge representation from downstream reasoning. In both paradigms, the knowledge store is constructed independently of the reasoning process, leaving no mechanism for reasoning failures to inform and improve the knowledge base. Addressing this gap requires a framework in which knowledge construction and reasoning are jointly optimized.

To this end, we propose \method{}, a \textit{co-evolving} framework that jointly learns to construct a structured knowledge base and reason over it, allowing downstream QA outcomes to directly shape what knowledge is constructed. Rather than treating KB construction as a fixed preprocessing step, our method treats it as a learnable component trained end-to-end alongside the reasoning module using QA outcome rewards (Figure~\ref{fig:overview_top}). 
Our method consists of two phases: (1) the model reads a set of related documents and constructs a structured KB that can be extracted and indexed offline for reuse across downstream queries; (2) at inference time, the model answers questions by retrieving and reasoning over the constructed KB. Both phases are optimized jointly, allowing reasoning signals to propagate back and guide the KB toward structures better aligned with compositional factual reasoning.

Our contributions are: (1) a co-evolving training framework that jointly optimizes KB construction and reasoning using outcome rewards, offering a scalable path to KB construction beyond the constraints of human or LLM annotation (Sec.~\ref{sec: method}); 
(2) competitive performance on knowledge-intensive QA benchmarks while enabling direct, training-free knowledge editing (Sec.~\ref{sec: main_results} and Sec.~\ref{sec:knowledge_editing});
and (3) evidence that co-evolution produces more complete and effective KBs and improves grounded reasoning
(Sec~\ref{sec: analysis}).

\input{section/sec_related_work}

\input{section/sec_method}

\input{table/algo}

\input{section/sec_exp}

\input{section/sec_ablation_study}

\section{Limitation}
\label{sec:limitation}

Our framework highlights a few challenges for future work. First, it relies on SFT for warm-starting, although scaling to larger base models with stronger prompting may reduce this dependence. 
Second, because QA rewards directly supervise only the subset of triplets retrieved during training, they do not guarantee faithfulness across the full constructed KB. While our analyses suggest that unsupported triplets primarily introduce retrieval noise rather than drive the observed gains, broader reward coverage and explicit verification remain important directions for future work.
Third, although our analysis in Sec~\ref{sec: analysis} suggests reward hacking is not the dominant behavior, the framework may still be vulnerable to it. 
Finally, a promising direction for future work is to introduce self-play or synthetic question generation with stronger verification, which may further strengthen the co-evolution of knowledge construction and reasoning.

\section{Conclusion}
We presented a co-evolving framework for knowledge-intensive question answering that jointly optimizes KB construction and reasoning via outcome-based reward signals. Rather than treating KB construction as a fixed preprocessing step, our method learns to construct structured knowledge representations end-to-end alongside the reasoning module, allowing reasoning failures to directly inform and improve the KB. At inference time, offline KB indexing decouples document processing from query time, offering a scalable path to KB construction beyond the constraints of human or LLM annotation.
Across the main QA benchmarks, \method{}-SFT performs comparably to IRCoT, while GRPO substantially improves over SFT, achieving comparable performance with Search-R1. \method{}-GRPO achieves the best results on MuSiQue and 2Wiki at both model scales, demonstrating that co-evolution can provide competitive reasoning performance while retaining a structured, inspectable, and reusable knowledge base.
Our work builds the foundation for this co-evolution, and leveraging these learned KBs to improve model capabilities beyond QA is a promising direction for future research.

\section{Acknowledgments}

This material is based on work supported by the AI Research Institutes program supported by the NSF and Intel Corporation under NSF award DMR-2433348. This research was also supported by the NSF under awards IIS-2505098, IIS-2530143 and OAC-2311521; a gift to the LinkedIn–Cornell Bowers Strategic Partnership; Gemini credits grant from Google. 
DG is supported by an Empire AI Postdoctoral Fellowship.
Opinions, findings and conclusions or recommendations expressed in this material are those of the
author(s) and do not necessarily reflect the views of the National Science Foundation. 
We thank the members of PIs' labs for helpful discussions.

\bibliography{colm2026_conference}
\bibliographystyle{colm2026_conference}

\clearpage
\appendix

\input{section/sec_appendix}
\end{document}

%% file: section/sec_related_work.tex
\section{Related Work}
\paragraph{Retrieval-Augmented LLMs and Search-Based Knowledge Agents.}
A growing body of work augments language models with external knowledge to improve factual reasoning. 
Early retrieval-augmented generation (RAG) methods retrieve relevant passages from an external corpus and condition generation on the retrieved context~\citep{lewis2020retrieval, izacard2021leveraging, guu2020retrieval, borgeaud2022improving, ram2023context}. 
More recent work moves beyond one-shot retrieval and studies \emph{search-based knowledge agents} that interleave reasoning with multi-turn information access.
Methods such as ReAct and Toolformer equip models with explicit tool-use capabilities, enabling them to decide when and how to invoke external tools during reasoning~\citep{schick2023toolformer, yao2022react}. 
Building on this direction, recent reinforcement learning approaches such as Search-R1 train models to issue search queries and interact with retrieval systems during step-by-step reasoning using only outcome supervision, leading to stronger multi-hop and knowledge-intensive reasoning~\citep{jin2025search}. 
Subsequent work further scales this paradigm to more realistic and long-horizon settings, including open-web research environments and broad knowledge-agent benchmarks~\citep{zheng2025deepresearcher, chen2025learning, chang2026karl}. 
While these methods substantially improve \emph{access} to external knowledge, they primarily operate over \emph{unstructured} information and provide limited control over the form, consistency, and editability of the knowledge being used.

\paragraph{Knowledge Storage, Compression, and Controllability in LLMs.}
A fundamental question in language modeling is how factual knowledge should be stored and represented. 
The dominant paradigm in modern LLMs is \emph{parametric} knowledge storage, where factual knowledge is implicitly compressed into model weights through next-token prediction~\citep{devlin2019bert, petroni2019language}. 
While this representation is compact and broadly generalizable, it is also inherently lossy: even large models often struggle to reliably retain long-tail facts and may hallucinate when knowledge is missing or weakly encoded~\citep{allen2023physics, kandpal2023large}. 
To address this, prior work has explored \emph{non-parametric} knowledge, including external corpora, knowledge databases, and learned memory that store knowledge externally~\citep{zhao2025pre, pouransari2025pretraining, bi2026parameters}.

A second challenge is that knowledge stored in model parameters is often highly entangled. 
Rather than being cleanly localized, factual information is distributed across shared representations together with linguistic patterns and other facts, a phenomenon often described as \emph{knowledge superposition}~\citep{elhage2022toy}. 
This makes knowledge difficult to inspect, edit, update, or remove, creating challenges for continual learning, knowledge editing and unlearning, and interpretability. 
Our work is motivated by these limitations. 
We study a setting in which knowledge is represented externally in a structured form that is more controllable and editable, while still being dynamically constructed and used by the model during reasoning.

\paragraph{Structured Knowledge Base Construction and Reasoning.}
\input{table/tab_kg_related_work}
As shown in Table~\ref{tab:related_work1}, prior KG construction methods such as EDC~\citep{zhang2024extract}, Wikontic~\citep{chepurova2026wikontic}, and AutoSchemaKG~\citep{bai2025autoschemakg} treat knowledge construction as a standalone pipeline, focusing on extraction, schema induction, canonicalization, or ontology-based filtering. Wikidata~\citep{vrandevcic2014wikidata} is a large human-curated knowledge base rather than an automated KG construction method. In contrast, our approach is schema-free and directly optimized with downstream QA supervision. As a result, we evaluate not only downstream QA performance but also the correctness, structure, and coverage of the constructed database.
Prior work on multi-hop reasoning over knowledge graphs largely assumes a fixed graph and improves inference over it, for example through path construction and pruning~\citep{tan2025paths}, LLM-based query planning~\citep{chen2024llm}, improved query representations~\citep{kim2024improving}, or traversal strategies such as hierarchical reinforcement learning~\citep{wang2025walk}. In contrast, our method jointly optimizes graph construction and reasoning during training, while keeping the graph fixed at inference time.

%% file: table/tab_kg_related_work.tex
\begin{table}[!htb]
\centering
\vspace{-1em}
\resizebox{\linewidth}{!}{
\begin{tabular}{lccccc}
\toprule
\textbf{Method} & \textbf{Construction} & \textbf{Schema} & \textbf{Quality Control} & \textbf{Downstream-Coupled} & \textbf{Persistence} \\
\midrule
EDC & Extract--Define--Canonicalize & Induced & Canonicalization & \ding{55} & Persistent \\
Wikontic & Multi-stage pipeline & Wikidata ontology & Ontology filtering + dedup & \ding{55} & Persistent \\
AutoSchemaKG & Autonomous pipeline & Induced & Schema-guided filtering & \ding{55} & Persistent \\
\rowcolor{lightgray} Ours & Joint RL training & Emergent (task-driven) & QA reward signal & \ding{51} & Persistent + Editable \\
\bottomrule
\end{tabular}
}
\caption{Comparison of knowledge graph construction methods.}
\label{tab:related_work1}
\vspace{-1em}
\end{table}

%% file: section/sec_method.tex
\section{Methodology}
\label{sec: method}
\subsection{Co-Evolving Framework}
We focus on \emph{entity-level atomic factual knowledge} as the basic unit of knowledge representation, where facts are stored as structured \texttt{(entity, relation, value)} triples. 

\begin{figure}[t!]
\captionsetup{skip=5pt}
\centering
\includegraphics[width=\linewidth]
{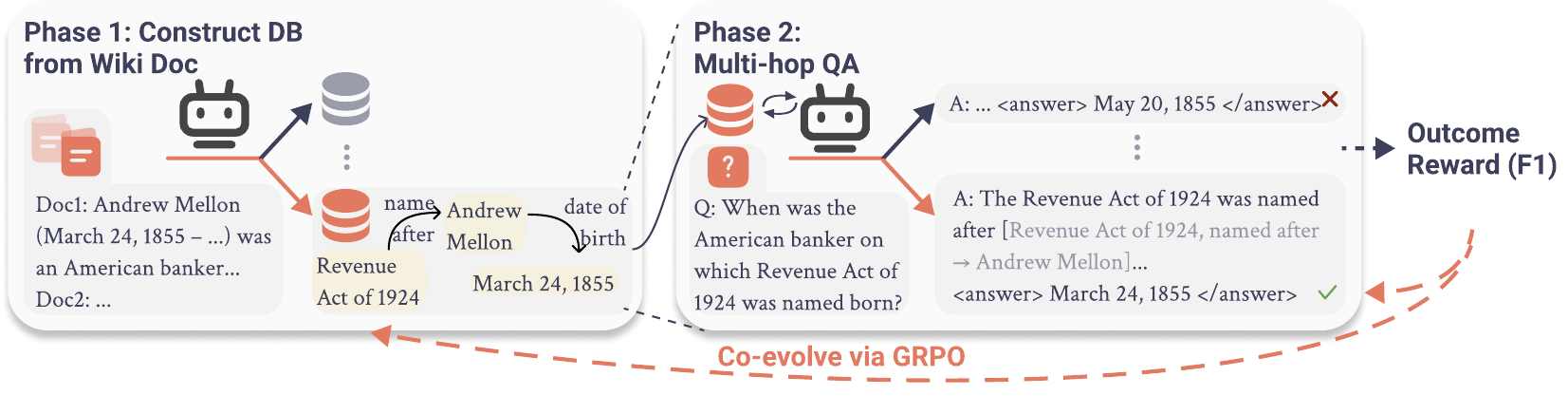}
\caption{\textbf{Overview of our co-evolving framework.} The model first constructs a structured knowledge base from context, then answers multihop questions by retrieving and reasoning over it. Both phases are jointly optimized so that knowledge construction and reasoning improve together.}
\label{fig:overview}
\vspace{-1em}
\end{figure}

An overview of our framework is illustrated in
Figure~\ref{fig:overview}.
We frame the problem as jointly training a single policy
$\pi_\theta$ that operates in two phases: \textit{knowledge base
construction} and \textit{question answering retrieving over the knowledge base}. Both phases share the same model parameters $\theta$,
allowing improvements in one phase to transfer to the other.

\paragraph{Phase 1: Knowledge Base Construction.}
Given a supporting passage $c$ for question $q$, the model
extracts a set of factual triplets
$\{(\texttt{entity}, \texttt{relation}, \texttt{value})\}$,
each representing an atomic fact. The collection of all extracted
triplets naturally forms a knowledge base
$G = (V, E)$, where the nodes $V$ are entities and values,
and the edges $E$ are relations.

\paragraph{Phase 2: Question Answering Retrieving over KB.}
To answer a factual question $q$, our model has access to the knowledge base $G$ through targeted retrieval. The model is able to issue tool calls, querying \texttt{(entity, relation)} pairs, and retrieving the corresponding value. Here is an example of Phase 2:

\begin{promptbox}
\texttt{The football club of Lionel Messi is <|db\_entity|> Lionel Messi <|db\_relationship|> Football club <|db\_return|> Inter Miami FC <|db\_end|> Inter Miami FC.\\<answer> Inter Miami FC </answer>}
\end{promptbox}
Here, \texttt{Inter Miami FC} is injected from the KB into the context at inference time. More details about retrieval in Appendix~\ref{app:model_arc}.

\subsection{GRPO Training}
We train the model using GRPO~\citep{shao2024deepseekmath}, with the full procedure summarized in Algorithm~\ref{alg:joint_train}.

% \ya{I recommend referencing specific line numbers in the text describing Alg1}

\paragraph{Sampling and Reward.}
For each question, we first sample $K$ candidate knowledge bases in Phase~1 (line~\ref{alg:line:kb}). For each constructed knowledge base, we then generate $M$ answer rollouts in Phase~2 (line~\ref{alg:line:qa}), yielding a total of $K \times M$ QA rollouts per question. We use F1 score between the predicted answer and ground truth answer as the Phase~2 reward (line~\ref{alg:line:qa_reward}). The reward for each Phase~1 knowledge base is defined as the average downstream QA reward over its associated Phase~2 rollouts (line~\ref{alg:line:kb_reward}).

\paragraph{Optimization.}
We compute group-relative advantages separately for the two phases (lines~\ref{alg:line:qa_adv}--\ref{alg:line:kb_adv}). For Phase 2, we normalize QA rewards within each (question, constructed KB) pair across its M answer rollouts. For Phase 1, we average Phase 2 rewards per KB and then normalize across the K candidate KBs for the same question. Unlike standard GRPO, where all rollouts in a group share the same input, our Phase 2 rollouts are conditioned on different constructed knowledge bases. 
% Normalizing across all $M$ rollouts therefore allows the model to compare reasoning quality. 
We then optimize $\pi_\theta$ using the standard GRPO clipped surrogate objective over the entire batch of Phase~1 and Phase~2 rollouts (line~\ref{alg:line:update}). To balance the gradient contribution of the two phases, we scale each Phase~1 advantage $A^{\text{kb}}_{b,k}$ by $M$, the number of associated Phase~2 rollouts. The full objective is provided in Appendix~\ref{app:grpo_objective}.
In Phase~2, retrieved result tokens are masked out from the loss.

The two phases co-evolve
during training: better knowledge bases enable better
answers, and sharper answer-level reward signals in turn
drive the construction of more effective knowledge bases.
We ablate this coupling in Sec~\ref{sec:coevolution-ablation}.

\subsection{Inference}
Given a collection of input documents, we first construct and index a structured KB using Phase 1. The resulting KB is then fixed and reused across downstream queries, where the model answers questions by querying the indexed KB through targeted retrieval.

%% file: table/algo.tex
\begin{algorithm}[t]
\small
\caption{Co-evolving KB Construction and QA via GRPO}
\label{alg:joint_train}
\begin{algorithmic}[1]
\Require Training set $\mathcal{D}$, batch size $B$, KB rollouts per sample $K$, QA rollouts per KB $M$, KB construction prompt $p^{\text{kb}}$, QA rollout prompt $p^{\text{qa}}$
\For{$t = 1$ to $T$}
    \State Sample minibatch of $B$ samples $\{(c_b, q_b, a_b^*)\}_{b=1}^B \sim \mathcal{D}$
    \Statex \hspace{\algorithmicindent}\textcolor{phaseblue}{\textbf{Phase 1: KB Construction}}
    % \State Construct $K$ knowledge bases per passage:
    \State For each passage $c_b$, construct $K$ knowledge bases: \label{alg:line:kb}
    \[
    \{G_{b,k}\}_{b=1,k=1}^{B,K} \sim \pi_\theta(\cdot \mid p^{\text{kb}}(c_b)) \quad \forall\, b
    \]
    \Statex \hspace{\algorithmicindent}\textcolor{phasegreen}{\textbf{Phase 2: QA Rollouts (conditioned on Phase 1)}}
    \State For each constructed knowledge base $G_{b,k}$, generate $M$ answers: \label{alg:line:qa}
    \[
    \{\hat{y}_{b,k,m}\}_{m=1}^{M} \sim \pi_\theta(\cdot \mid p^{\text{qa}}(q_b), G_{b,k}) \quad \forall\, b,k
    \]
    \Statex \hspace{\algorithmicindent}\textcolor{phaseorange}{\textbf{Reward and Advantage Computation}}
    \State Compute QA rewards for $\hat{y}_{b,k,m}$: \label{alg:line:qa_reward}
    $r_{b,k,m} \gets r(\hat{y}_{b,k,m},\, a_b^*) \quad \forall\, b,k,m$
    \State Compute KB rewards for $G_{b,k}$: $\bar{r}_{b,k} \gets \frac{1}{M}\sum_{m=1}^{M} r_{b,k,m} \quad \forall\, b,k$ \label{alg:line:kb_reward}
    \State Compute QA advantages $\{A^{\text{qa}}_{b,k,m}\}$ by normalizing $\{r_{b,k,m}\}_{m=1}^{M}$ within each $b, k$ \label{alg:line:qa_adv}
    \State Compute KB advantages $\{A^{\text{kb}}_{b,k}\}$ by normalizing $\{\bar{r}_{b,k}\}_{k=1}^K$ within each $b$ \label{alg:line:kb_adv}
    \Statex \hspace{\algorithmicindent}\textcolor{phasered}{\textbf{Joint GRPO Update}}
    \State Update $\pi_\theta$ with a joint GRPO step over the concatenated Phase~1 and Phase~2 rollout batch, \label{alg:line:update}
    \Statex \hspace{\algorithmicindent}using advantages $\{A^{\text{kb}}_{b,k}\}$ and $\{A^{\text{qa}}_{b,k,m}\}$
\EndFor
\end{algorithmic}
\end{algorithm}

%% file: section/sec_exp.tex
\section{Experimental Setup}
\label{sec:exp_setup}
\paragraph{Training Setting.} 
We use \qwens{} and \qwenm{}~\citep{yang2025qwen3} and train on 7k HotpotQA~\citep{yang2018hotpotqa} examples. We generate 6k SFT trajectories from HotpotQA using Gemini-2.5-Flash~\citep{comanici2025gemini}, covering both Phase~1 KB construction and Phase~2 question answering, and fine-tune each model for 3 epochs. We subsequently train with GRPO on 7k HotpotQA examples for 500 steps using an F1-based outcome reward.
For retrieval, we use \texttt{all-MiniLM-L6-v2}~\citep{reimers-gurevych-2019-sentence} with a similarity threshold of 0.6 (returning \texttt{unknown} otherwise) and top-$k=4$. 
Additional details are provided in Appendix~\ref{app:model_arc}.

\paragraph{Baselines.}
We compare \method{} against several representative baselines:
\begin{itemize}[topsep=2pt, itemsep=2pt, parsep=2pt, leftmargin=*]
    \item \textit{Direct}: direct answer generation without external retrieval.
    \item \textit{RAG}~\citep{lewis2020retrieval}: retrieval-augmented generation over text chunks.
    \item \textit{IRCoT}~\citep{trivedi2023interleaving}: interleaves chain-of-thought reasoning with iterative BM25 retrieval. We run IRCoT over the same benchmark-specific corpus used by the other retrieval methods.
    \item \textit{Search-R1}~\citep{jin2025search}: trains a model to interleave
multi-turn search with reasoning over unstructured text. Since Search-R1
was not released for \textsc{Qwen3}, we reimplement it with matched training
data, training steps, and overall training budget. In our \textsc{Qwen3}
reimplementation, the original prompt often reverted to long-form reasoning
without search; we therefore provide a single in-context example illustrating
targeted search and the expected tool-call format. Full implementation and
prompt details are provided in Appendix~\ref{app:searchr1}.
\end{itemize}
We refer to the model after supervised fine-tuning as \method{}-SFT, and to the final model after GRPO post-training as \method{}-GRPO.

\paragraph{Benchmarks and Metrics.}
We evaluate on three multi-hop question answering benchmarks: HotpotQA (7,405 samples)~\citep{yang2018hotpotqa}, MuSiQue (2,417 samples)~\citep{trivedi2022musique}, and 2WikiMultiHopQA (12,576 samples)~\citep{ho2020constructing}. We additionally evaluate on PopQA (1,399 samples) for single-hop factual QA.
Since training uses HotpotQA, we treat HotpotQA as in-domain and the
remaining benchmarks as out-of-domain evaluation.
We report exact match (EM) as the primary metric.

\paragraph{Retrieval and Knowledge Base Setup.}

For each benchmark, all retrieval-based methods use the same source documents. Text-based baselines index document chunks for retrieval, while \method{} constructs a structured KB from the same documents and retrieves over the extracted triplets. We aggregate all chunks or triplets within each benchmark into a single retrieval datastore.

\section{Results}

\subsection{Knowledge Intensive QA}
\label{sec: main_results}

\input{table/results_multihop_benchmark}

We report results on three multi-hop QA benchmarks and PopQA in
Table~\ref{tab:main_results}. Across both model scales and all benchmarks,
\method{}-GRPO consistently improves over \method{}-SFT, increasing average
EM by 5.1 points at 1.7B and 9.8 points at 4B. This consistent gain shows that
downstream QA reward provides an effective learning signal for jointly improving
knowledge construction and multi-hop reasoning.

\textsc{Search-R1} provides our closest comparison to the full method: both approaches
use outcome-based RL to learn retrieval and reasoning, but \textsc{Search-R1} retrieves
directly from unstructured text, whereas \method{} learns to construct and reason
over a structured KB. Under our matched Qwen3 training setup, \method{}-GRPO
achieves comparable overall performance to \textsc{Search-R1} (41.3 vs.\ 42.4 average
EM at 1.7B and 46.6 vs.\ 49.0 at 4B), with complementary strengths across
benchmarks. 
These
results suggest that replacing unstructured text search with a structured
knowledge interface does not substantially sacrifice the effectiveness of
RL-trained retrieval and reasoning.

The two approaches differ, however, in what they provide beyond QA accuracy.
Structured knowledge represents individual facts explicitly, allowing the
knowledge available to the model to be inspected, edited, and controlled.
As shown in Table~\ref{fig:confiqa_results}, this structure also enables targeted
knowledge updates that the model can effectively incorporate during reasoning.
Thus, \method{} achieves competitive RL-trained reasoning performance while
retaining the controllability and reusability that motivate structured external
knowledge.

\subsection{Reasoning with Edited External Knowledge}
\label{sec:knowledge_editing}
\input{table/table_confiqa}

Structured knowledge allows individual facts to be directly edited.
We evaluate whether models can effectively use such updates for multi-hop reasoning on ConFiQA-MR~\citep{bi2024context}, which introduces counterfactual facts that conflict with a model's parametric knowledge. We consider three settings: MR-ORIG uses the original facts, while MR-CF-100 and MR-CF-356 replace 100 and 356 examples with conflict-free counterfactual versions and update the corresponding knowledge stores.

As shown in Figure~\ref{fig:confiqa_results}, \method{} remains substantially stronger than Search-R1 after these knowledge edits across both model scales.
These results show that its structured KB can be directly updated and effectively used for reasoning without retraining.

\subsection{Analysis: Knowledge Base Coverage and Grounded Reasoning}
\label{sec: analysis}

One of the main outcomes of our method is that it enables an affordable small model to \emph{construct} its own knowledge base rather than relying on a fixed external one. A natural concern, however, is that GRPO provides supervision only through downstream QA outcomes, meaning that only a subset of the constructed knowledge may receive direct learning signal. We therefore conduct a comprehensive analysis of both the \emph{quality of the constructed knowledge base} and the model’s \emph{ability to reason over it}.

\begin{figure}[t]
  \centering
  \vspace{-0.5em}

  \begin{subfigure}[t]{0.32\linewidth}
    \centering
    \includegraphics[width=\linewidth]{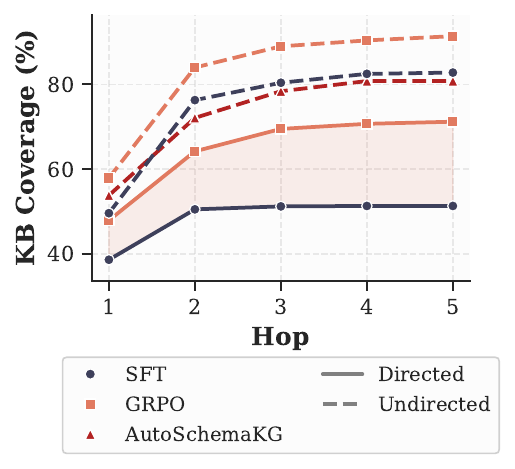}
    \caption{Knowledge Base Coverage.}
    \label{fig:db_coverage}
  \end{subfigure}
  \hfill
  \begin{subfigure}[t]{0.32\linewidth}
    \centering
    \includegraphics[width=\linewidth]{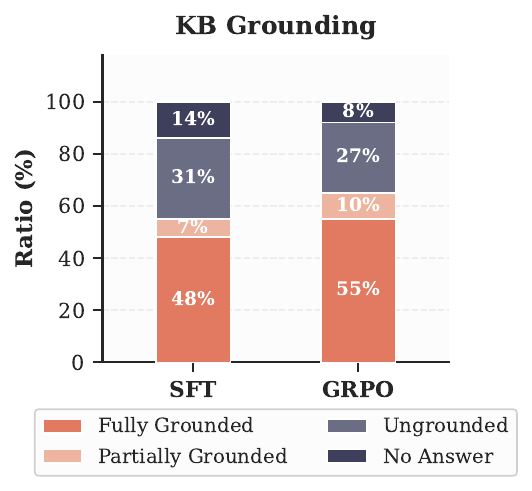}
    \caption{Knowledge Grounding.}
    \label{fig:db_grounding}
  \end{subfigure}
  \hfill
  \begin{subfigure}[t]{0.32\linewidth}
    \centering
    \includegraphics[width=\linewidth]{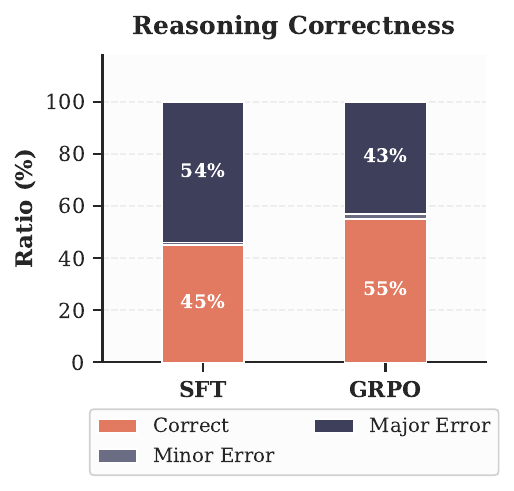}
    \caption{Reasoning composition.}
    \label{fig:reasoning_stacked}
  \end{subfigure}

\caption{
Co-evolution improves KB coverage and grounded reasoning.
(a) GRPO improves answer reachability across hop depths and outperforms static KB construction methods.
(b--c) GRPO also improves answer grounding and reasoning correctness over the constructed KB.
The analysis is conducted on \qwenm{} using 1K HotpotQA examples.}
  \label{fig:fig_db_quality}
  \vspace{-1em}
\end{figure}

\textbf{Co-evolution improves the constructed knowledge base.}
Figure~\ref{fig:db_coverage} measures whether the answer is reachable from question entities
within a bounded number of hops in the constructed KB. 

GRPO consistently improves reachability over SFT across
hop depths, suggesting that outcome supervision encourages the construction of knowledge
paths that better support multi-hop reasoning.

\textbf{Co-evolution improves reasoning over the knowledge base.}
The gains are not limited to KB construction. Using Gemini-2.5-Flash as a judge, we evaluate whether the
model’s answers are supported by retrieved
knowledge base entries and whether its
reasoning chains are logically correct. As shown in
Figures~\ref{fig:db_grounding} and ~\ref{fig:reasoning_stacked}, \method{} increases the fraction of answers fully grounded
in retrieved knowledge from 48\% to 55\%, while reasoning correctness increases from
45\% to 55\%. These results together indicate that co-evolution improves both the
knowledge available to the model and its ability to use that knowledge during reasoning.
Additional analyses of KB structure, lookup behavior, and
faithfulness are provided in Appendix~\ref{sec:additional_analysis}.

%% file: table/results_multihop_benchmark.tex
\begin{table}[t]
\centering
\resizebox{0.8\linewidth}{!}{
\begin{tabular}{lcccc|c}
\toprule
Method
& HotpotQA$^*$
& MuSiQue
& 2Wiki
& PopQA
& Avg \\
\midrule

\textbf{\qwens} & & & & & \\

Direct
& 12.1
& 0.7
& 21.7
& 16.0
& 12.6 \\

RAG
& 25.4
& 6.3
& 21.8
& 57.8
& 27.8 \\

IRCoT
& 23.7
& 12.2
& 44.0
& 60.7
& 35.2 \\

Search-R1$^\dagger$
& 44.1
& 14.3
& 45.9
& 65.3
& 42.4 \\

\rowcolor{lightgray}
\method{}-SFT
& 29.5
& 13.3
& 44.9
& 57.0
& 36.2 \\

\rowcolor{midgray}
\method{}-GRPO
& 37.3
& 18.6
& 47.8
& 61.4
& 41.3 \\

\midrule

\textbf{\qwenm} & & & & & \\

Direct
& 12.3
& 1.5
& 18.5
& 13.7
& 11.5 \\

RAG
& 32.4
& 6.7
& 20.3
& 63.1
& 30.6 \\

IRCoT
& 33.0
& 19.7
& 27.1
& 65.0
& 36.2 \\

Search-R1$^\dagger$
& 51.1
& 22.6
& 50.4
& 71.8
& 49.0 \\

\rowcolor{lightgray}
\method{}-SFT
& 35.6
& 16.7
& 40.7
& 54.3
& 36.8 \\

\rowcolor{midgray}
\method{}-GRPO
& 46.1
& 26.4
& 51.4
& 62.5
& 46.6 \\

\bottomrule
\end{tabular}
}
\caption{
Evaluation on multi-hop QA benchmarks and PopQA (EM).
$^*$ marks the in-domain benchmark.
$^\dagger$~Reimplemented on \textsc{Qwen3} backbone.
See Section ~\ref{sec:exp_setup}.
}
\label{tab:main_results}
\vspace{-0.5em}
\end{table}

%% file: table/table_confiqa.tex
\begin{wrapfigure}{r}{0.52\linewidth}
    \centering
    \vspace{-1em}
    \includegraphics[width=\linewidth]{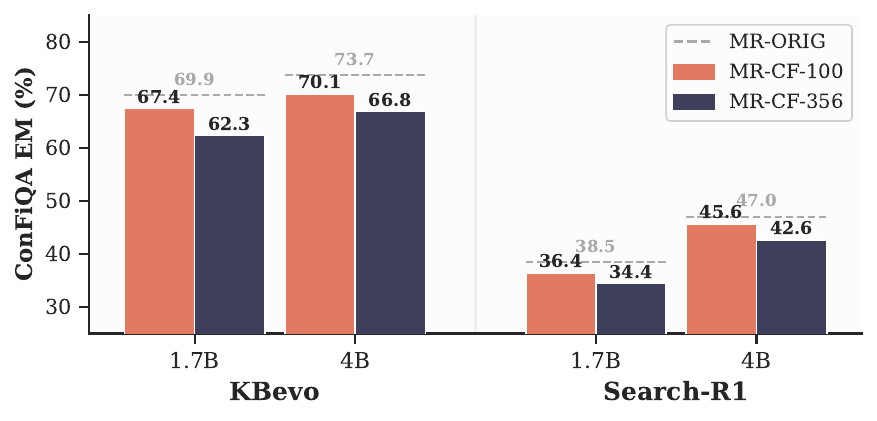}
    \vspace{-1em}
    \caption{\textbf{Reasoning with edited external knowledge on ConFiQA-MR (EM).}
    % Exact-match accuracy under the original and two counterfactual knowledge settings.
    \method{} maintains strong performance under counterfactual knowledge edits, showing that its structured KB can be directly updated and effectively used for reasoning without model retraining.
}
    \label{fig:confiqa_results}
    \vspace{-1em}
\end{wrapfigure}

%% file: section/sec_ablation_study.tex
\subsection{What Does Co-Evolution Buy Us?}
\label{sec:coevolution-ablation}

\input{table/ablation_1phase}

Our framework jointly optimizes KB construction and reasoning, making their contributions intertwined. We therefore ablate this coupling in two complementary ways: 
(a) swapping the KB at inference while keeping the reasoning policy fixed, which isolates the quality of the constructed KB; and (b) fixing the KB throughout training, which removes co-evolution between knowledge construction and reasoning.

\paragraph{(a) Swap KB at inference.}
Holding the \method{}-1.7B GRPO reasoning policy fixed, our learned KB substantially outperforms KBs constructed by existing static pipelines: 35.6 average EM compared with 26.2 for EDC~\citep{zhang2024extract} and 24.6 for AutoSchemaKG~\citep{bai2025autoschemakg}. This is despite both baselines using substantially larger extraction models. Replacing our KB with a Gemini-constructed KB yields 37.2 average EM, only 1.6 points higher overall, suggesting that our learned KB captures much of the downstream utility of the stronger teacher KB.

\paragraph{(b) Without co-evolution.}
We next keep an externally constructed KB fixed throughout training and optimize only the reasoning policy. With AutoSchemaKG, performance drops from 35.6 to 28.5 average EM, showing that reasoning optimization alone cannot compensate for a weaker fixed KB. A fixed Gemini KB reaches 36.1 average EM, comparable to the full co-evolving model, providing a strong reference point for fixed-KB training with a frontier-model teacher.

\subsection{Is Supervised Fine-Tuning Necessary Before Reinforcement Learning?}

\begin{figure}[t]
  \centering
  \vspace{-0.5em}

  \begin{subfigure}[t]{0.28\linewidth}
    \centering
    \includegraphics[width=\linewidth]{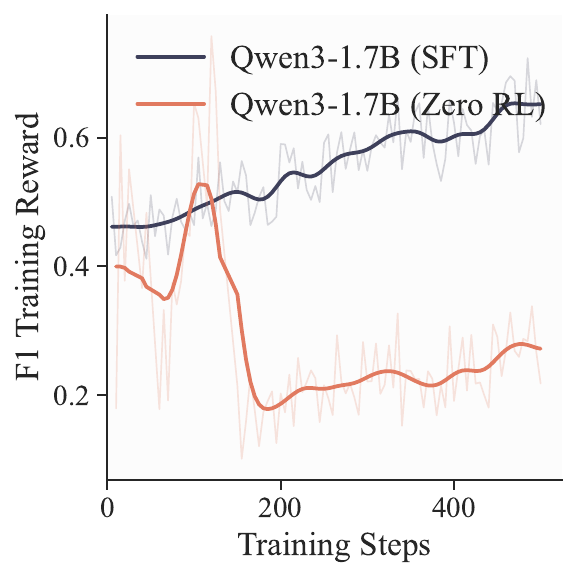}
    \caption{Zero-RL reward collapse.}
    \label{fig:zero_rl_reward}
  \end{subfigure}
  \hfill
  \begin{subfigure}[t]{0.28\linewidth}
    \centering
    \includegraphics[width=\linewidth]{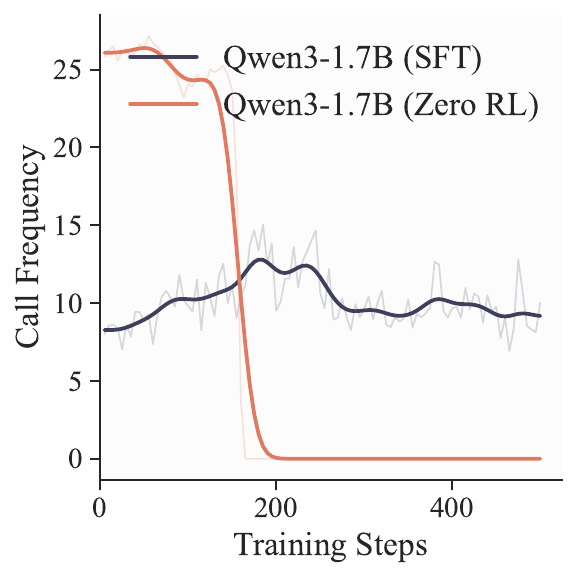}
    \caption{Lookup usage drops.}
    \label{fig:zero_rl_call_frequency}
  \end{subfigure}
  \hfill
  \begin{subfigure}[t]{0.28\linewidth}
    \centering
    \includegraphics[width=\linewidth]{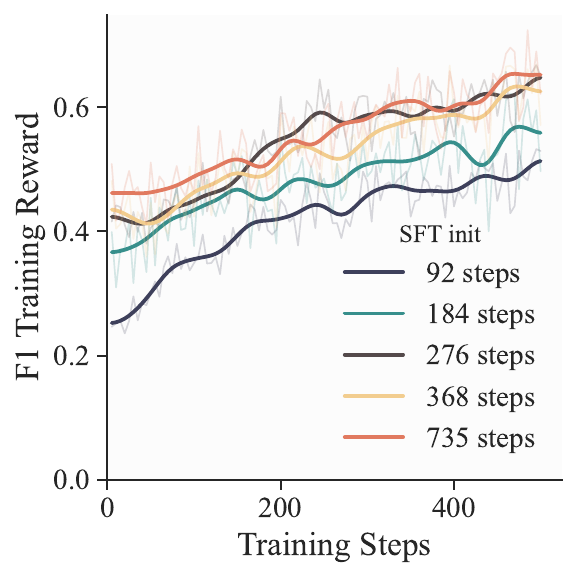}
    \caption{Early SFT is sufficient.}
    \label{fig:early_sft_reward}
  \end{subfigure}
% \vspace{-1em}
  \caption{
  Effect of SFT warmup before RL. 
  Starting RL directly from the base model leads to collapse in both reward and lookup usage, while even relatively early SFT checkpoints are sufficient to enter the structured reasoning regime and achieve similar improvement during RL.
  }
  \label{fig:sft_warmup}
  \vspace{-1em}
\end{figure}

We study whether the model can be trained directly with RL from the base model, optionally with format rewards, following~\citet{zeng2025simplerl}. In practice, we find that this is insufficient: even with dedicated prompt and format rewards, the model fails to reliably perform database construction and reasoning over it. 
In Figure~\ref{fig:zero_rl_call_frequency}, the model struggles to produce valid structured traces and collapses to relying on
internal parametric knowledge instead of external lookup.
This suggests that structured reasoning with dblookup is substantially out-of-distribution
for the base model, even with dedicated prompt tuning, making SFT a crucial initialization step. This aligns with prior work showing that RL requires an established capability foundation to be effective \citep{gandhi2025cognitive, yue2025does}.
% \lx{Why RL is still useful after SFT}

We further study the role of SFT by initializing RL from checkpoints at different stages of SFT (Figure~\ref{fig:early_sft_reward}). Earlier SFT checkpoints start with lower performance but quickly improve during RL and eventually reach comparable final performance. This suggests that even a relatively early SFT stage is sufficient to bootstrap the model into the structured reasoning regime, after which RL can effectively refine both database construction and downstream reasoning. A more thorough investigation of this trade-off is left for future work.

\begin{figure}[htbp]
    \centering
    \includegraphics[width=\textwidth]{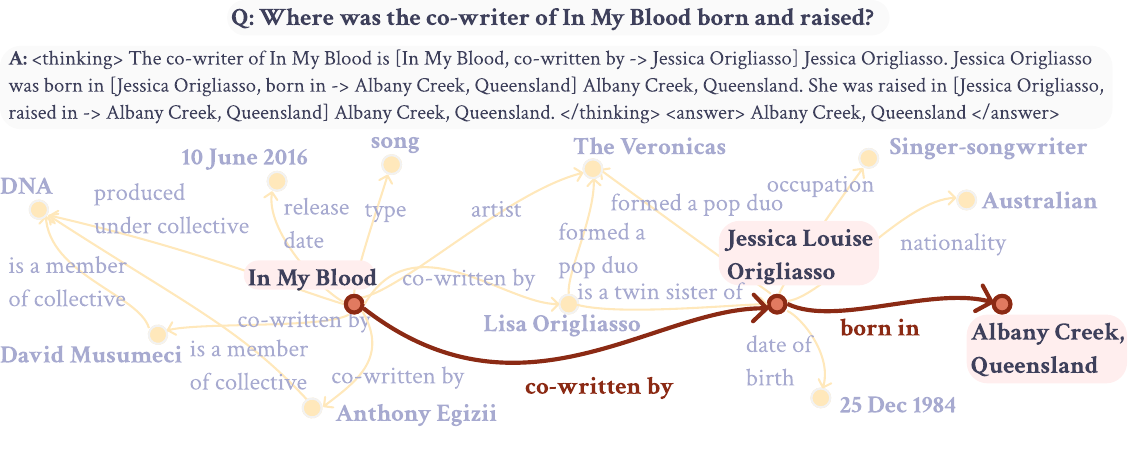}
    \caption{A representative example. The model constructs a knowledge base from source documents and then reasons over it to answer a multi-hop question.}
    \label{fig:kg_case_study}
    \vspace{-1em}
\end{figure}

%% file: table/ablation_1phase.tex
\begin{table}[t]
\centering
\begin{tabular}{lccc|c}
\toprule
Method & HotpotQA$^*$ & MuSiQue & 2Wiki & Avg \\
\midrule

\multicolumn{5}{l}{\textit{(a) Swap KB at inference (reasoning policy fixed)}} \\
\rowcolor{gray!15}
\method{}-1.7B GRPO$^\dagger$
    & 40.3 & 19.3 & 47.3 & 35.6 \\
\quad -- EDC DB (Mistral-7B)
    & 30.2 & 8.3 & 40.1 & 26.2 \\
\quad -- AutoSchemaKG DB (Llama-3.1-8B)
    & 29.0 & 16.1 & 28.8 & 24.6 \\
\quad -- Gemini DB
    & 38.1 & 19.0 & 54.5 & 37.2 \\

\midrule

\multicolumn{5}{l}{\textit{(b) Train with a fixed KB (w/o Co-evolution)}} \\
\rowcolor{gray!15}
\method{}-1.7B GRPO$^\dagger$
    & 40.3 & 19.3 & 47.3 & 35.6 \\
\quad -- AutoSchemaKG DB (Llama-3.1-8B)
    & 34.1 & 13.5 & 38.0 & 28.5 \\
\quad -- Gemini DB
    & 38.6 & 15.1 & 54.5 & 36.1 \\

\bottomrule
\end{tabular}

\caption{
Ablating co-evolution. Our learned KB substantially outperforms KBs from static construction methods, while training with a fixed KB generally underperforms joint co-evolution. Results are EM on 1K examples per dataset due to the cost of Gemini-based KB construction. \textsuperscript{$\dagger$} \method{} results are evaluated on this same 1K subset and therefore differ slightly from the full-set results in Table~\ref{tab:main_results}.
$^*$ marks the in-domain benchmark.}
\label{tab:ablation_coevolution}
\end{table}

%% file: section/sec_appendix.tex
\input{appendix/model_architecture_and_training_details}

\input{appendix/llm_as_a_judge_prompt}

\clearpage
\section{Detailed Analysis}
\label{app:detailed_analysis}

\paragraph{Checkpoint note.}
Analyses in Section~\ref{sec: analysis} and Appendix~\ref{app:detailed_analysis} use an earlier checkpoint trained with retrieval top-$k=1$, while the main results use the final $k=4$ configuration. Absolute QA results in these analyses are therefore not directly comparable to Table~\ref{tab:main_results}.

\input{appendix/additional_analysis}
\input{appendix/rebuttal_analysis}

%% file: appendix/model_architecture_and_training_details.tex
\section{Method Details}

\subsection{GRPO objective function}
\label{app:grpo_objective}
The joint GRPO objective is:
\begin{equation}
    \mathcal{J}(\theta) = \lambda \, \mathcal{J}^{\text{kb}}(\theta) + \mathcal{J}^{\text{qa}}(\theta)
\end{equation}
where $\lambda$ balances the contribution of the two phases.
Since each question produces $K$ KB trajectories but
$K \times M$ QA trajectories, we set $\lambda = M$ to
match the total weight of both phases.

For Phase 1 (KB construction):
\begin{equation}
\begin{aligned}
\mathcal{J}^{\text{kb}}(\theta)
&=
\mathbb{E}_{\substack{
(c,q,a^*) \sim \mathcal{D}, \\
\{G_k\}_{k=1}^K \sim
\pi_{\theta_{\text{old}}}(\cdot \mid p^{\text{kb}}(c))
}}
\Bigg[
\sum_{k=1}^{K} \frac{1}{|G_k|}
\sum_{t=1}^{|G_k|}
\\
&\qquad\qquad
\min\!\Big(
\rho^{\text{kb}}_{k,t}\, A^{\text{kb}}_k,\;
\text{clip}\big(
\rho^{\text{kb}}_{k,t},\, 1-\epsilon,\, 1+\epsilon
\big)\,
A^{\text{kb}}_k
\Big)
\Bigg].
\end{aligned}
\end{equation}

For Phase 2 (QA), conditioned on the constructed KB $G_k$:
\begin{equation}
\begin{aligned}
\mathcal{J}^{\text{qa}}(\theta)
&=
\mathbb{E}_{\substack{
(c,q,a^*) \sim \mathcal{D},\;
G_k \sim \pi_{\theta_{\text{old}}}(\cdot \mid p^{\text{kb}}(c)) \\
\{\hat{y}_{k,m}\}_{m=1}^M \sim
\pi_{\theta_{\text{old}}}(\cdot \mid p^{\text{qa}}(q, G_k))
}}
\Bigg[
\sum_{k=1}^{K} \sum_{m=1}^{M}
\frac{1}{|\hat{y}_{k,m}|}
\sum_{t=1}^{|\hat{y}_{k,m}|}
\\
&\qquad\qquad
\min\!\Big(
\rho^{\text{qa}}_{k,m,t}\, A^{\text{qa}}_{k,m},\;
\text{clip}\big(
\rho^{\text{qa}}_{k,m,t},\, 1-\epsilon,\, 1+\epsilon
\big)\,
A^{\text{qa}}_{k,m}
\Big)
\Bigg].
\end{aligned}
\end{equation}

where the importance ratios are:
\begin{align}
    \rho^{\text{kb}}_{k,t} &= \frac{\pi_\theta(G_{k,t} \mid
        p^{\text{kb}}(c),\, G_{k,<t})}
        {\pi_{\theta_{\text{old}}}(G_{k,t} \mid
        p^{\text{kb}}(c),\, G_{k,<t})} \\[6pt]
    \rho^{\text{qa}}_{k,m,t} &= \frac{\pi_\theta(\hat{y}_{k,m,t}
        \mid p^{\text{qa}}(q, G_k),\, \hat{y}_{k,m,<t})}
        {\pi_{\theta_{\text{old}}}(\hat{y}_{k,m,t} \mid
        p^{\text{qa}}(q, G_k),\, \hat{y}_{k,m,<t})}
\end{align}

\textbf{Advantage calculation.}
For Phase~1 (KB construction), we first aggregate QA rewards into a
KB-level reward $\bar{r}_{b,k} = \frac{1}{M}\sum_{m=1}^{M} r_{b,k,m}$,
then normalize across the $K$ KBs within each question $b$:
\begin{equation}
    A^{\text{kb}}_{b,k} = \frac{\bar{r}_{b,k} - \mu^{\text{kb}}_b}
    {\sigma^{\text{kb}}_b}, \quad
    \mu^{\text{kb}}_b = \frac{1}{K}\sum_{k=1}^{K} \bar{r}_{b,k}
\end{equation}

For Phase~2 (QA), we normalize rewards within each $(b, k)$ group
across the $M$ rollouts:
\begin{equation}
    A^{\text{qa}}_{b,k,m} = \frac{r_{b,k,m} - \mu^{\text{qa}}_{b,k}}
    {\sigma^{\text{qa}}_{b,k}}, \quad
    \mu^{\text{qa}}_{b,k} = \frac{1}{M}\sum_{m=1}^{M} r_{b,k,m}
\end{equation}

Both baselines are derived solely from the QA reward signal, but at different aggregation scopes: Phase-1 baselines average across the K candidate KBs for a question, while Phase-2 baselines average across the M rollouts conditioned on a single KB.

\subsection{Training and Inference Cost}
\label{app:cost}

The SFT warm-start is light: 6k traces, with a 0.4-epoch checkpoint already sufficient
(Figure~\ref{fig:early_sft_reward}). GRPO uses more rollouts per question, but each is short, a
lookup injects roughly ten tokens, whereas document-based search appends a full passage every turn
and grows the context across turns~\citep{jin2025search}, which is why
Table~\ref{tab:main_results} matches rollout budget rather than rollout count. At inference the
knowledge base is built offline once per document and reused, so the per-query cost is one Phase-2
generation plus embedding lookups, with no $K\!\times\!M$ sampling, no teacher model, and no schema
induction stage.
\input{table/appendix_cost}

\clearpage
\section{Experimental Setup}

\subsection{Training Details}
\label{app:model_arc}
\subsubsection{SFT Data Generation}
\label{app:sft_data_curation}
We generate synthetic two-phase supervision using Gemini 2.5 Flash. For each training example, the teacher model first produces a Phase 1 knowledge base constructed from the supporting context, and then generates a Phase 2 reasoning trace that answers the question by issuing database lookups over the constructed knowledge base. These trajectories are used as supervised fine-tuning data.

\begin{promptbox}
Extract triplets from the following context. Each triplet must be of the form (entity, relationship, value).

Rules:

\begin{itemize}
    \item An entity is the subject of a sentence. Do not use pronouns (e.g., ``he'' or ``it'') to describe an entity; use the full descriptive name.

    \item Relationships refer to any characteristic of an entity. Even if a relationship is not explicitly mentioned, you must include it. Relationships can describe things the entity was involved in or any descriptive characteristic of that entity. If an entity is described with multiple characteristics at once, create separate triplet entries for each characteristic.
\end{itemize}

Example: From the text ``Albert Einstein was a German theoretical physicist best known for developing the theory of relativity,'' the triplets are
(Albert Einstein, nationality, Germany),
(Albert Einstein, occupation, theoretical physicist), and
(Albert Einstein, best known for, developing the theory of relativity).

For example, from the text ``In 1796, Napoleon commanded a military campaign against the Austrians and their Italian allies in the War of the First Coalition,'' you would extract:
(Napoleon, commanded a military campaign in 1796 against, the Austrians and their Italian allies in the War of the First Coalition),
(Napoleon, commanded a military campaign against the Austrians and their Italian allies in the War of the First Coalition, in 1796), and
(Napoleon, commanded a military campaign against Austrians and Italians, in the War of the First Coalition in 1796).

Given these examples, generate explicit and implicit triplets from the following context.

Rules:

\begin{itemize}
    \item The relationships should be as precise as possible while not requiring external information to look them up.

    \item Do NOT include multiple (entity, relation) pairs with different values. In such cases, the relationship or entity is not precise enough, and you must add more precision.
\end{itemize}

Here is the context:

\{context\}
\end{promptbox}

\subsubsection{Database and Retrieval Implementation}

The database is a collection of \texttt{(entity, relationship, value)} triplets. Each \texttt{(entity, relationship)} pair is embedded using \texttt{sentence-transformers/all-MiniLM-L6-v2}, and during retrieval, the \texttt{(entity, relationship)} query is matched to the nearest pairs via cosine similarity of the embeddings, retrieving the top k=4 results above a 0.6
cosine similarity threshold. We found that this combination of threshold and top-k
yielded the highest performance after SFT training. Furthermore, we augment the database with \textit{reverse-index entries}: for each \texttt{(entity, relationship, value)} triplet, we include \texttt{(value, relationship, entity)} in the database. We find that this improves initial SFT performance, the learning ability of the model during reinforcement learning, and the final performance of the model.
% \lx{Or anyone, can you check it}

\subsubsection{Training hyperparameters}

We train \method{} in two stages. 
In the first stage, we perform supervised fine-tuning (SFT) from \qwens{} and \qwenm{} on Gemini-generated two-phase reasoning traces constructed from HotpotQA. 
We train for 3 epochs with learning rate $5\times10^{-5}$, per-device batch size 24, gradient accumulation 2, cosine learning rate decay, warmup ratio 0.1, weight decay 0.01, and maximum sequence length 2048. 

In the second stage, we initialize GRPO from the SFT checkpoint and continue training on HotpotQA for 500 steps with learning rate $5\times10^{-6}$, effective batch size 512, cosine learning rate decay, warmup ratio 0.1, and maximum gradient norm 1.0. 
For each question, we sample $K{=}4$ database (KB) rollouts and $N{=}32$ QA rollouts ($M{=}8$ QA rollouts per KB candidate, $36$ total rollouts), jointly optimizing database construction and reasoning with an outcome-based F1 reward. We use top-$p$ sampling ($p=0.95$), temperature 1.0, top-$k=4$, retrieval threshold 0.6, maximum completion length 1024, and vLLM colocation with gradient checkpointing for efficient batched inference.

\input{table/appendix_hyperparameter}

\input{appendix/searchr1_reimplementation}
\input{appendix/confiqa_cf_setup}

%% file: table/appendix_cost.tex
\begin{table}[H]
\centering
\label{tab:cost_breakdown}
\resizebox{\linewidth}{!}{
\begin{tabular}{lll}
\toprule
Phase & Cost & Recurs at inference? \\
\midrule
\multicolumn{3}{l}{\textit{Training (one-time)}} \\
\quad SFT warm-start & 6k HotpotQA traces (Gemini 2.5 Flash), 3 epochs & No \\
\quad GRPO & 36 rollouts/question (4 KB + 4$\times$8 QA), 500 steps, $\beta=0$ & No \\
\midrule
\multicolumn{3}{l}{\textit{Inference (per query)}} \\
\quad KB construction & Built offline once per document, reused across queries & Amortized \\
\quad Retrieval & top-$k=4$ MiniLM lookup, no LLM call & Per lookup \\
\quad QA & One short Phase-2 generation & Per query \\
\bottomrule
\end{tabular}
}
\caption{Cost breakdown of \method{}. The additional expense is confined to a one-time SFT warm-start and the $K + K\!\times\!M$ rollouts during GRPO. At inference the knowledge base is constructed offline once per document and reused across queries, so per-query cost is a single Phase-2 generation plus embedding lookups; no $K\!\times\!M$ sampling and no teacher model recur at inference.}
\end{table}

%% file: table/appendix_hyperparameter.tex
\begin{table*}[t]
\centering
\resizebox{0.92\linewidth}{!}{
\begin{tabular}{lll}
\toprule
\textbf{Category} & \textbf{SFT} & \textbf{GRPO} \\
\midrule
\textbf{Model \& Data} \\
Base model
& Qwen3-1.7B / Qwen3-4B
& SFT checkpoint (step 368 for 1.7B, step 735 for 4B) \\
Dataset
& Gemini 2-phase trajectories (HotpotQA, 6k)
& HotpotQA \\
Train size / Eval size
& 6k / --
& 7000 / 100 \\
Use special DB lookup tokens
& Yes
& Yes \\
\midrule
\textbf{Optimization} \\
Training objective
& Supervised fine-tuning
& GRPO \\
Learning rate
& $5\times10^{-5}$
& $5\times10^{-6}$ \\
Epochs / Max steps
& 3 epochs
& 500 steps \\
Scheduler
& Cosine
& Cosine \\
Warmup ratio
& 0.1
& 0.1 \\
Weight decay
& 0.01
& -- \\
Max grad norm
& --
& 1.0 \\
$\beta$
& --
& 0.0 \\
\midrule
\textbf{Batching} \\
Per-device train batch size
& 24 / 8
& 16 / 8 \\
Per-device eval batch size
& 24 / 8
& 32 \\
Gradient accumulation
& 2 / 6
& 8 / 16 \\
Effective batch size
& 48
& 512 \\
Num GPUs
& 1
& 4 (B200) \\
\midrule
\textbf{Sequence / Generation} \\
Max sequence length
& 2048 / 1024
& 4096 (vLLM), 1024 completion \\
Top-$p$ / Temperature / Top-$k$
& --
& 0.95 / 1.0 / 4 \\
Num generations (train / eval)
& --
& $36 = 4 + 4{\times}8$ (4 KB + 32 QA) \\
\midrule
\textbf{Two-phase \& Retrieval} \\
Two-phase training
& Implicit in supervision data
& Enabled \\
Num DB rollouts
& --
& 4 \\
QA rollouts per DB
& --
& 8 \\
Reward function
& --
& F1 \\
Phase-1 prompt type
& --
& SFT \\
Phase-1 DB weight mode
& --
& Count \\
Retrieval threshold / top-$k$
& --
& 0.6 / 4 \\
Use inverses
& --
& Enabled \\
Adaptive $k$
& --
& Disabled \\
\midrule
\textbf{System} \\
Precision
& bf16
& bf16 / vLLM \\
Gradient checkpointing
& --
& Enabled \\
vLLM (colocate)
& --
& Enabled \\
vLLM memory utilization
& --
& 0.4 / 0.15 \\
\midrule
\textbf{Logging \& Checkpointing} \\
Logging steps
& 10
& 5 \\
Eval strategy
& Epoch
& Steps (every 100) \\
Save strategy
& Steps
& Steps \\
Save steps
& 0.125 epoch
& 25 \\
Save total limit
& 8
& 5 \\
\bottomrule
\end{tabular}
}
\caption{Training configurations for \method{} on multi-hop QA. We first perform supervised fine-tuning (SFT) on two-phase KB-construction and QA trajectories, then continue joint optimization with GRPO.}
\label{tab:train_config_all}
\vspace{-0.5em}
\end{table*}

%% file: appendix/searchr1_reimplementation.tex
\subsection{Search-R1 Reimplementation}
\label{app:searchr1}

\paragraph{Training and retrieval setup.}
Since Search-R1~\citep{jin2025search} does not release checkpoints for
\textsc{Qwen3}, we reimplement it on \qwens{} and \qwenm{} using
\texttt{verl} with multi-turn tool interaction. We initialize directly from
the base models, without SFT, following the original Search-R1 training
paradigm. To enable a controlled comparison with \method{}, we use the same
7K HotpotQA training examples, train for 500 update steps, and use the same
token-level F1 outcome reward. Each training step samples 16 prompts with
5 rollouts per prompt. We use temperature $1.0$, top-$p=0.95$, and
top-$k=4$, with at most five assistant turns and a maximum response length
of 2,048 tokens.

Search-R1 retrieves over unstructured text using an E5-base-v2 dense
retriever with a FAISS index, returning the top three passages for each
search query. Retrieved tool responses are truncated to 1,024 characters.
The search tool is exposed through the native \textsc{Qwen3} tool-calling
interface as a function taking a list of search queries.

\paragraph{Prompt adaptation for \textsc{Qwen3}.}
In preliminary runs, directly applying the original Search-R1 prompt to
\textsc{Qwen3} frequently resulted in long-form chain-of-thought generation
without invoking the search tool. We therefore make two prompt-level
adaptations while leaving the training and retrieval setup unchanged.
First, we use \texttt{<thinking>...</thinking>} rather than
\texttt{<think>...</think>} to avoid interfering with \textsc{Qwen3}'s
native thinking interface. Second, we provide a single three-hop
in-context demonstration illustrating concise decomposition, iterative
search, and the exact JSON tool-call format expected by the
\textsc{Qwen3} tool interface. No supervised trajectories are used for
Search-R1.

\begin{promptbox}
\textbf{System:} You are a helpful and harmless assistant.

\textbf{User:} Answer the given question. You must conduct reasoning inside \texttt{<thinking>} and \texttt{</thinking>} first every time you get new information. Each \texttt{<thinking>} should be ONE BRIEF SENTENCE stating what you need next --- not detailed analysis. After reasoning, if you lack knowledge, call the search tool. When you have enough information, provide the answer inside \texttt{<answer>} and \texttt{</answer>} without detailed illustrations.

Here is an example showing the expected format and brevity:

Question: What is the population of the capital city of the country where the inventor of the World Wide Web was born?

\texttt{<thinking>}I need to identify the inventor of the World Wide Web and where he was born.\texttt{</thinking>}

\texttt{<tool\_call>}\{"name": "search", "arguments": \{"query\_list": ["inventor of the World Wide Web birthplace"]\}\}\texttt{</tool\_call>}

\texttt{<tool\_response>}Doc 1 (Title: "Tim Berners-Lee"): Sir Timothy John Berners-Lee, also known as TimBL, is an English computer scientist best known as the inventor of the World Wide Web. He was born in London, England.\texttt{</tool\_response>}

\texttt{<thinking>}Tim Berners-Lee was born in England, so I need the capital of the United Kingdom.\texttt{</thinking>}

\texttt{<tool\_call>}\{"name": "search", "arguments": \{"query\_list": ["capital of the United Kingdom"]\}\}\texttt{</tool\_call>}

\texttt{<tool\_response>}Doc 1 (Title: "London"): London is the capital and largest city of England and the United Kingdom.\texttt{</tool\_response>}

\texttt{<thinking>}The capital is London, so I need its population.\texttt{</thinking>}

\texttt{<tool\_call>}\{"name": "search", "arguments": \{"query\_list": ["population of London"]\}\}\texttt{</tool\_call>}

\texttt{<tool\_response>}Doc 1 (Title: "London"): London has a population of approximately 9 million people.\texttt{</tool\_response>}

\texttt{<thinking>}The population is about 9 million.\texttt{</thinking>}

\texttt{<answer>}9 million\texttt{</answer>}

Now answer this question: \texttt{\{QUESTION\}}
\end{promptbox}

%% file: appendix/confiqa_cf_setup.tex
\subsection{ConFiQA Evaluation Details}
\label{app:confiqa-evaluation}

\paragraph{Data and counterfactual conditions.}
We use the ConFiQA-MR evaluation set released with
Context-DPO~\citep{bi2024context}, which provides paired original and
counterfactual contexts, labeled reasoning paths represented as
\texttt{(entity, relation, value)} triplets, answers, and aliases. We first
shuffle the dataset once with seed 42 and then retain the first 1,000 examples.
We use the same 1,000 source examples and ordering in every condition.

We construct three evaluation sets. MR-ORIG uses the original context, labeled
reasoning path, answer, and aliases for every example. To construct the
counterfactual conditions, we define a forward knowledge key as
\texttt{(entity, relation)} and require that the aggregate labeled paths contain
at most one distinct value for every such key. We solve this selection problem
with a deterministic binary program and use lexicographic tie-breaking that
prefers counterfactual substitutions at earlier positions in the fixed
ordering. MR-CF-100 uses an conflict-free selection of 100
counterfactual examples. MR-CF-356 uses 356 counterfactual examples, the
maximum feasible number under this forward-key definition. All
remaining examples retain their original versions.

\paragraph{Database and corpus construction.}
For \method{}, we aggregate the labeled triplets from the reasoning paths using
the selected version of each example into a single database and use that
database for all questions in the corresponding setting. The model does not
receive the gold path for the current question directly; it retrieves from the
shared database constructed from all 1,000 examples. 

For Search-R1~\citep{jin2025search}, we instead build the retrieval corpus and
E5-base-v2 index from the corresponding 1,000 ConFiQA contexts directly. Each
passage is the full natural-language context supplied by ConFiQA, which embeds
% the labeled reasoning path in longer descriptive text. We evaluate the corresponding 1.7B and 4B merged checkpoints using
% their ICL three-hop tool-call format, with temperature 1.0, top-$p$ 0.95,
% sampling top-$k$ 4, and at most five turns. Following the checkpoint's
% training-time configuration, each retrieval response is truncated to its first 1,024 characters.
the labeled reasoning path in longer descriptive text. Search-R1 retrieves the
top three passages from this closed corpus. We
evaluate using the ICL
three-hop tool-call format, with temperature 1.0, top-$p$ 0.95, sampling
top-$k$ 4, at most five turns, and a maximum response length of 2,048 tokens.
Following the checkpoints' training-time configuration, each combined retrieval
response is truncated to its first 1,024 characters.

\paragraph{Metric.}
We report normalized exact match against the
condition-specific answer and its aliases. Before comparison, predictions and
reference answers are lowercased, stripped of punctuation and the articles
\emph{a}, \emph{an}, and \emph{the}, and normalized for whitespace.
\method{} uses the checkpoints and inference configuration described in
Appendix~\ref{app:model_arc}.

%% file: appendix/llm_as_a_judge_prompt.tex
\subsection{LLM as a judge prompt}
\label{sec:judge_prompts}
LLM as a judge prompt for evaluating knowledge base faithfulness and soundness.
\begin{promptbox}
\textbf{[System]}\\
You are an expert evaluator for knowledge graph databases. Your task is to evaluate the quality of a knowledge database (a set of triplets) that was generated from a given source context (a wiki document).

You will evaluate the database on two dimensions: \textbf{Faithfulness} and \textbf{Soundness}.

\vspace{0.5em} \hrule \vspace{0.5em}

\textbf{Dimension 1: Faithfulness (Is the triplet grounded in the context?)}

For each triplet (entity, relation, value), determine whether the information it expresses can be found in or directly inferred from the provided context.
\begin{enumerate}
    \item \textbf{Faithful}: The triplet's meaning is supported by the context. It does not need to be a verbatim match — reasonable paraphrasing and direct inference are acceptable. Inference must be strictly logical based ONLY on the provided text, without incorporating external world knowledge.
    \item \textbf{Hallucinated}: The triplet contains information that does NOT appear in and cannot be directly inferred from the context. This includes fabricated entities, invented relations, values not mentioned or implied in the source, or facts that are only true based on external knowledge not present in the context.
\end{enumerate}

\vspace{0.5em} \hrule \vspace{0.5em}

\textbf{Dimension 2: Soundness (Is the triplet well-formed and useful?)}

For each triplet that is NOT hallucinated, check for the following soundness issues. A single triplet may have multiple issues.

\textbf{Issue types:}
\begin{enumerate}
    \item \textbf{Ambiguous Entity/Value}: The entity or value uses unclear references (e.g., pronouns like "he", "it", "the company") instead of explicit names, or is ambiguous without additional context.
    \begin{itemize}
        \item Bad: (He, founded, the company)
        \item Good: (Elon Musk, founded, SpaceX)
    \end{itemize}
   \item \textbf{Trivial / Meaningless}: The triplet expresses an obvious, near-tautological, or extremely low-information fact that adds no meaningful knowledge.
   \begin{itemize}
       \item Bad: (Einstein, is, a person) — states the obvious, no useful information
       \item Good: (Einstein, developed, Theory of Relativity)
   \end{itemize}
   \item \textbf{Non-specific}: Any field (entity, relation, or value) is vague or imprecise where the context provides more specific information.
   \begin{itemize}
       \item Bad: (Company X, revenue, very high) — when the context states an exact figure
       \item Good: (Company X, revenue, $574.8B$)
   \end{itemize}
   \item \textbf{Malformed / Span Error}: The triplet has severe structural problems. Only flag this when the relation or entity field contains a full clause or sentence (with multiple verbs or subordinate clauses), when entity fields pack multiple distinct entities, or when the triplet is incomplete / unparseable. Short prepositional phrases in the relation field are acceptable.
   \begin{itemize}
       \item Bad: (Apple Inc., released the first iPhone which revolutionized the market in, 2007) — relation is an entire clause
       \item Good: (Apple Inc., released, the first iPhone)
   \end{itemize}
   \item \textbf{Reversed / Misplaced Roles}: The entity and value are swapped, or the directionality of the relation is incorrect, causing the triplet to express the opposite or a nonsensical meaning.
   \begin{itemize}
       \item Bad: (Theory of Relativity, developed, Einstein) — subject and object are reversed
       \item Good: (Einstein, developed, Theory of Relativity)
   \end{itemize}
\end{enumerate}

\vspace{0.5em} \hrule \vspace{0.5em}

\textbf{Output Format}

\begin{enumerate}
    \item `reasoning`: Brief explanation covering why the triplet is faithful/hallucinated, and if faithful, why specific soundness issues were flagged (if any). Keep it concise.
    \item For hallucinated triplets, skip soundness evaluation — set soundness\_issues to [].
    \item Evaluate every triplet in the database. Do not skip any.
\end{enumerate}
\end{promptbox}

\begin{promptbox}
\textbf{[User]}\\
\textbf{Context (Source Wiki Document):}
\{context\}

\textbf{Knowledge Database to Evaluate:}
\{database\_triplets\}

Evaluate each triplet in this database according to the rubric. Return your evaluation as JSON.
\end{promptbox}

LLM as a judge prompt for evaluating knowledge base grounding and reasoning correctness.

\begin{promptbox}
\textbf{[System]}\\
You are an expert evaluator for knowledge-graph-augmented question answering. You will be given:
1. A question
2. A knowledge database (a set of triplets)
3. A model's completion (its thinking process and DB lookups)

You will evaluate the completion on two dimensions: \textbf{DB Grounding} and \textbf{Reasoning Correctness}.

\vspace{0.5em} \hrule \vspace{0.5em}

\textbf{Dimension 1: DB Grounding (Is the final answer based on DB lookups?)}

Determine whether the model's final answer is supported by the information retrieved from DB lookups.
\begin{itemize}
    \item \textbf{fully\_grounded}: The final answer is entirely based on information retrieved from DB lookups. Every key claim in the answer can be traced back to a DB return value.
    \item \textbf{partially\_grounded}: The final answer is partly based on DB lookups and partly based on the model's own internal knowledge. Some claims are supported by DB returns, but others are not.
    \item \textbf{ungrounded}: The final answer has no basis in the DB lookups. The model either used entirely internal knowledge, or fabricated an answer that contradicts or ignores what the DB returned.
    \item \textbf{no\_answer}: The model failed to produce a final answer (e.g., got stuck in a loop, gave up, or the completion was cut off before answering).
\end{itemize}

\vspace{0.5em} \hrule \vspace{0.5em}

\textbf{Dimension 2: Reasoning Correctness (Is the multi-hop reasoning chain correct?)}

Evaluate whether the model's reasoning process — how it chains together information from multiple DB lookups to arrive at a conclusion — is logically sound.
\begin{itemize}
    \item \textbf{correct}: The reasoning chain is logically valid. Each step follows from the previous one, and the conclusion is properly supported by the retrieved information.
    \item \textbf{minor\_error}: The overall reasoning direction is correct, but there are small issues — e.g., a skipped step that doesn't affect the final answer, or a slightly imprecise inference.
    \item \textbf{major\_error}: The reasoning chain has significant logical errors — e.g., drawing wrong conclusions from the retrieved data, confusing entities, or making invalid logical jumps.
\end{itemize}

Note: If the grounding is "no\_answer", set reasoning to "major\_error" since no valid reasoning was completed.

\vspace{0.5em} \hrule \vspace{0.5em}

\textbf{Output Instructions}
\begin{itemize}
    \item `grounding\_reasoning`: Brief explanation of why the grounding label was chosen.
    \item `reasoning\_reasoning`: Brief explanation of why the reasoning correctness label was chosen.
    \item Evaluate based on the FINAL answer, not intermediate thoughts.
    \item The DB lookup format in completions is:\texttt{\textless{}|db\_entity|\textgreater{}ENTITY \textless{}|db\_relationship|\textgreater{}RELATIONSHIP\textless{}|db\_return|\textgreater{}VALUE\textless{}|db\_end|\textgreater{}}
\end{itemize}

\end{promptbox}

\begin{promptbox}
\textbf{[User]}\\
\textbf{Question}:
\{question\}

\textbf{Knowledge Database}:
\{database\_triplets\}

\textbf{Model Completion}:
\{completion\}

Evaluate this completion according to the rubric.
\end{promptbox}

%% file: appendix/additional_analysis.tex
\subsection{Intrinsic Quality of the Constructed Knowledge Base}
\label{sec:additional_analysis}

We further analyze how co-evolution changes the constructed knowledge base,
independently of downstream QA.

\paragraph{Co-evolution improves KB structure and coverage.}
As shown in Table~\ref{tab:db_quality}, GRPO produces larger, better-connected
KBs and reduces failed or redundant lookups. Compared with static KG
construction methods, \method{} also achieves higher triplet quality and
substantially better multi-hop answer reachability
(Table~\ref{tab:static_kg_quality}). These results suggest that outcome
supervision encourages knowledge structures that are better suited for
downstream reasoning.

\input{table/analysis_kg_n_lookup}
\input{table/appendix_static_kg_quality}

\paragraph{Improved coverage does not imply uniformly higher faithfulness.}
GRPO increases coverage and connectivity, but also increases hallucination and
soundness-issue rates relative to SFT
(Table~\ref{tab:phase1_results}). We therefore examine below whether these
unfaithful triplets contribute to the downstream gains.

\input{table/appendix_db_correctness}

%% file: table/analysis_kg_n_lookup.tex
\begin{table}[H]
\centering
\small
\setlength{\tabcolsep}{6pt}
\begin{tabular}{l cccc}
\toprule
& \multicolumn{2}{c}{\textbf{Qwen3-1.7B}}
& \multicolumn{2}{c}{\textbf{Qwen3-4B}} \\
\cmidrule(lr){2-3} \cmidrule(lr){4-5}
\textbf{Metric} & SFT & GRPO & SFT & GRPO \\
\midrule

\textbf{KB Structure} & & & & \\
$|E|$ (unique entities)   & 156.6 & 169.4 & 156.8 & 186.6 \\
$|R|$ (unique relations)  & 111.3 & 135.9 & 109.7 & 175.3 \\
\#Triplets                & 166.2 & 225.9 & 165.7 & 299.1 \\
Avg Degree                & 2.15  & 2.71  & 2.14  & 3.30  \\
\#Components $\downarrow$ & 13.1  & 10.3  & 13.0  & 7.9   \\
Giant Comp.\ (\%) $\uparrow$ & 57.2  & 65.1  & 57.8  & 73.8  \\

\midrule
\textbf{Lookup Efficiency} & & & & \\
Redundancy Rate (\%) $\downarrow$ & 6.3 & 2.3 & 7.2 & 6.3 \\
Unknown Rate (\%) $\downarrow$    & 6.6 & 2.3 & 7.1 & 2.0 \\

\bottomrule
\end{tabular}
\caption{Knowledge base structure and lookup efficiency on HotpotQA\textsuperscript{$\dagger$}, averaged per example. \textit{Redundancy Rate} is the fraction of lookups repeating an entity--relation pair already queried; \textit{Unknown Rate} is the fraction matching no entry. Co-evolution yields larger, better-connected knowledge bases that the reasoning policy also queries more successfully. \textsuperscript{$\dagger$}These diagnostics were computed using an earlier set of model checkpoints and are retained for reference.}
\label{tab:db_quality}
\end{table}

%% file: table/appendix_static_kg_quality.tex
\begin{table}[hbt]
\centering
\resizebox{\linewidth}{!}{
\begin{tabular}{lccc|ccc|cc}
\toprule
& \multicolumn{3}{c|}{{Triplet Quality}}
& \multicolumn{3}{c|}{{Graph Structure}}
& \multicolumn{2}{c}{{Answer Reachability}} \\
\cmidrule(lr){2-4} \cmidrule(lr){5-7} \cmidrule(lr){8-9}
KB Source
& Prec.$\uparrow$ & Rec.$\uparrow$ & F1$\uparrow$
& $|R|$ & \#Comp.$\downarrow$ & Giant \%$\uparrow$
& $\leq$2 hop$\uparrow$ & $\leq$4 hop$\uparrow$ \\
\midrule

EDC (Mistral-7B)
& 0.864 & 0.477 & 0.600
& -- & -- & --
& 63.7\% & 69.5\% \\

AutoSchemaKG (Llama-3.1-8B)
& 0.888 & 0.838 & 0.860
& 58.8 & 18.8 & 51.0\%
& 72.1\% & 80.8\% \\

\rowcolor{lightgray}
\method{}-1.7B GRPO
& 0.925 & 0.923 & 0.924
& 135.9 & 10.3 & 65.1\%
& 81.3\% & 88.4\% \\

\rowcolor{midgray}
\method{}-4B GRPO
& {0.938} & {0.952} & {0.945}
& {175.3} & {7.9} & {73.8\%}
& {84.0\%} & {90.4\%} \\

\bottomrule
\end{tabular}
}
\caption{Intrinsic knowledge base quality on HotpotQA, independent of downstream QA.
\method{} produces more accurate and better-connected KBs than static construction pipelines,
with substantially higher multi-hop answer reachability.}
\label{tab:static_kg_quality}
\vspace{-0.5em}
\end{table}

%% file: table/appendix_db_correctness.tex
\begin{table}[th]
\centering
\resizebox{0.75\linewidth}{!}{
\begin{tabular}{lccccc}
\toprule
{Category}
  & {AutoSchemaKG}
  & \multicolumn{2}{c}{{Qwen3-1.7B}}
  & \multicolumn{2}{c}{{Qwen3-4B}} \\
\cmidrule(lr){3-4} \cmidrule(lr){5-6}
  & {(Llama-3.1-8B)} & {SFT} & {GRPO}
  & {SFT} & {GRPO} \\
\midrule
Valid Triplets $\uparrow$      & 35.6\% & 92.3\% & 85.0\% & 96.0\% & 84.3\% \\
Hallucinated $\downarrow$      & 11.0\% & 3.3\%  & 7.5\%  & 1.5\%  & 6.2\%  \\
Soundness Issues $\downarrow$  & 53.4\% & 4.4\%  & 7.4\%  & 2.6\%  & 9.5\%  \\
\bottomrule
\end{tabular}
}
\caption{Triplet-level correctness on HotpotQA (100 samples), judged with the same LLM-as-judge protocol across all knowledge bases (Appendix~\ref{sec:judge_prompts}). \textit{Valid Triplets} are free of hallucination and soundness issues. GRPO increases both hallucination and soundness-issue rates relative to SFT; we examine whether this explains the downstream gains in Appendix~\ref{app:faithfulness}.}
\label{tab:phase1_results}
\end{table}

%% file: appendix/rebuttal_analysis.tex
\subsection{Are the Gains Driven by Faithful Knowledge?}
\label{app:faithfulness}

Because QA reward directly supervises only retrieved triplets, we examine whether
unretrieved knowledge degrades and whether the QA gains rely on hallucinated content.

\paragraph{Unretrieved knowledge remains faithful.}
Although only about 5\% of constructed triplets are retrieved during reasoning,
unused triplets are nearly as faithful as used ones
(Table~\ref{tab:faithfulness_analysis}). This suggests that sparse reward exposure
does not substantially degrade the unretrieved portion of the KB.

\paragraph{Hallucinated knowledge does not drive the gains.}
Removing unfaithful triplets slightly improves EM
(Table~\ref{tab:faithfulness_analysis}), indicating that they primarily act as
retrieval noise. Moreover, GRPO substantially improves both EM and grounding even
when all retrieved triplets are faithful
(Table~\ref{tab:faithfulness_conditioned}). Thus, the gains are better explained by
improved use of faithful knowledge than by exploiting hallucinated content.

\input{table/appendix_faithfulness}
\input{table/appendix_faithfulness_conditioned}

%% file: table/appendix_faithfulness.tex
\begin{table}[hbt]
\centering
\resizebox{\linewidth}{!}{
\begin{tabular}{lcc|cccc}
\toprule
& \multicolumn{2}{c|}{\textbf{Post-hoc Filtering (EM)}}
& \multicolumn{4}{c}{\textbf{Faithfulness by Reward Exposure}} \\
\cmidrule(lr){2-3} \cmidrule(lr){4-7}
Model & No Filter & Filter Unfaithful
& Used Faithful\% & Used Cov.\% & Unused Faithful\% & Unused Cov.\% \\
\midrule
\method{}-1.7B SFT  & --   & --   & 97.6 & 5.3 & 96.8 & 94.7 \\
\rowcolor{lightgray}
\method{}-1.7B GRPO & 36.2 & 38.4 & 94.3 & 5.3 & 92.7 & 94.7 \\
\method{}-4B SFT    & --   & --   & 98.6 & 5.4 & 98.2 & 94.6 \\
\rowcolor{midgray}
\method{}-4B GRPO   & 41.2 & 43.6 & 94.1 & 5.1 & 93.2 & 94.9 \\
\bottomrule
\end{tabular}
}
\caption{Faithfulness analysis of the constructed knowledge base on HotpotQA (1K examples). \textit{Left:} removing triplets judged unfaithful slightly improves EM, indicating that hallucinated triplets act primarily as retrieval noise rather than an exploitable shortcut. \textit{Right:} triplets retrieved during reasoning (\textit{Used}) and never retrieved (\textit{Unused}) show similar faithfulness under the same evaluation protocol (Appendix~\ref{sec:judge_prompts}), suggesting that sparse reward exposure does not substantially degrade the unretrieved portion of the knowledge base.}
\label{tab:faithfulness_analysis}

\vspace{-0.5em}
\end{table}

%% file: table/appendix_faithfulness_conditioned.tex
\begin{table}[H]
\centering
\resizebox{\linewidth}{!}{
\begin{tabular}{llccccc}
\toprule
Model & Retrieval Faithfulness & Share\% & EM & Fully Grounded & Partially Grounded & Ungrounded \\
\midrule
\method{}-1.7B SFT  & All faithful     & 84.0 & 33.0 & 49.6 & 10.6 & 30.2 \\
\method{}-1.7B SFT  & Has hallucinated & 16.0 & 17.5 & 28.1 &  9.4 & 39.4 \\
\rowcolor{lightgray}
\method{}-1.7B GRPO & All faithful     & 60.7 & 40.3 & 59.7 & 11.1 & 27.1 \\
\rowcolor{lightgray}
\method{}-1.7B GRPO & Has hallucinated & 39.3 & 24.2 & 41.6 & 10.2 & 39.5 \\
\midrule
\method{}-4B SFT    & All faithful     & 90.7 & 35.0 & 50.1 &  8.8 & 27.2 \\
\method{}-4B SFT    & Has hallucinated &  9.3 & 24.7 & 35.5 &  7.5 & 35.5 \\
\rowcolor{midgray}
\method{}-4B GRPO   & All faithful     & 58.0 & 48.8 & 68.2 &  9.7 & 16.8 \\
\rowcolor{midgray}
\method{}-4B GRPO   & Has hallucinated & 42.0 & 30.5 & 42.2 & 10.0 & 27.7 \\
\bottomrule
\end{tabular}
}
\caption{Faithfulness-conditioned grounding and QA performance on HotpotQA (1K examples). Reasoning chains are partitioned by whether all retrieved triplets are judged faithful or at least one is hallucinated. GRPO substantially improves both EM and grounding even when reasoning relies entirely on faithful triplets, indicating that the gains are not driven by hallucinated content. The fraction of chains retrieving at least one hallucinated triplet nevertheless increases under GRPO.}
\label{tab:faithfulness_conditioned}
\end{table}

%% file: colm2026_conference.bib
@article{zeng2025simplerl,
  title = {Simplerl-zoo: Investigating and taming zero reinforcement learning for open base models in the wild},
  author = {Zeng, Weihao and Huang, Yuzhen and Liu, Qian and Liu, Wei and He, Keqing and Ma, Zejun and He, Junxian},
  journal = {arXiv.org},
  year = {2025},
  doi = {10.48550/arXiv.2503.18892},
}

@article{ram2023context,
  title = {In-context retrieval-augmented language models},
  author = {Ram, Ori and Levine, Yoav and Dalmedigos, Itay and Muhlgay, Dor and Shashua, Amnon and Leyton-Brown, Kevin and Shoham, Yoav},
  journal = {Transactions of the Association for Computational Linguistics},
  volume = {11},
  pages = {1316--1331},
  year = {2023},
  publisher = {MIT Press},
  doi = {10.1162/tacl_a_00605},
}

@inproceedings{guu2020retrieval,
  title = {Retrieval augmented language model pre-training},
  author = {Guu, Kelvin and Lee, Kenton and Tung, Zora and Pasupat, Panupong and Chang, Mingwei},
  booktitle = {International conference on machine learning},
  pages = {3929--3938},
  year = {2020},
  organization = {PMLR},
  journal = {International Conference on Machine Learning},
}

@article{bi2024context,
  title = {Context-DPO: Aligning Language Models for Context-Faithfulness},
  author = {Bi, Baolong and Huang, Shaohan and Wang, Yiwei and Yang, Tianchi and Zhang, Zihan and Huang, Haizhen and Mei, Lingrui and Fang, Junfeng and Li, Zehao and Wei, Furu and others},
  journal = {Findings of the Association for Computational Linguistics: ACL 2025},
  year = {2025},
  pages = {10280-10300},
  doi = {10.18653/v1/2025.findings-acl.536},
  publisher = {Association for Computational Linguistics},
}

@inproceedings{borgeaud2022improving,
  title = {Improving language models by retrieving from trillions of tokens},
  author = {Borgeaud, Sebastian and Mensch, Arthur and Hoffmann, Jordan and Cai, Trevor and Rutherford, Eliza and Millican, Katie and Driessche, George van den and Lespiau, Jean-Baptiste and Damoc, Bogdan and Clark, Aidan and others},
  booktitle = {International conference on machine learning},
  pages = {2206--2240},
  year = {2021},
  organization = {PMLR},
  journal = {International Conference on Machine Learning},
}

@inproceedings{zhao2025pre,
  title={Pre-training limited memory language models with internal and external knowledge},
  author={Zhao, Linxi and Zalouk, Sofian and Belardi, Christian and Lovelace, Justin and Zhou, Jin and Noonan, Ryan and Go, Dongyoung and Weinberger, Kilian and Artzi, Yoav and Sun, Jennifer},
  booktitle={International Conference on Learning Representations},
  volume={2026},
  pages={60117--60152},
  year={2026}
}

@article{elhage2022toy,
  title = {Toy models of superposition},
  author = {Elhage, Nelson and Hume, Tristan and Olsson, Catherine and Schiefer, Nicholas and Henighan, T. and Kravec, S. and Hatfield-Dodds, Zac and Lasenby, R. and Drain, Dawn and Chen, Carol and others},
  journal = {arXiv.org},
  year = {2022},
  doi = {10.48550/arXiv.2209.10652},
}

@article{bommasani2021opportunities,
  title={On the opportunities and risks of foundation models},
  author={Bommasani, Rishi and Hudson, Drew A and Adeli, Ehsan and Altman, Russ and Arora, Simran and von Arx, Sydney and Bernstein, Michael S and Bohg, Jeannette and Bosselut, Antoine and Brunskill, Emma and others},
  journal={arXiv preprint arXiv:2108.07258},
  year={2021}
}

@article{pouransari2025pretraining,
  title = {Pretraining with hierarchical memories: separating long-tail and common knowledge},
  author = {Pouransari, Hadi and Grangier, David and Thomas, C and Kirchhof, Michael and Tuzel, Oncel},
  journal = {arXiv.org},
  year = {2025},
  doi = {10.48550/arXiv.2510.02375},
}

@article{meng2022locating,
  title = {Locating and editing factual associations in gpt},
  author = {Meng, Kevin and Bau, David and Andonian, Alex and Belinkov, Yonatan},
  journal = {Neural Information Processing Systems},
  volume = {35},
  pages = {17359--17372},
  year = {2022},
  doi = {10.52202/068431-1262},
  publisher = {Neural Information Processing Systems Foundation, Inc. (NeurIPS)},
}

@inproceedings{wang2025walk,
  title = {Walk wisely on graph: Knowledge graph reasoning with dual agents via efficient guidance-exploration},
  author = {Wang, Zijian and Wang, Bin and Jing, Haifeng and Li, Huayu and Dou, Hongbo},
  booktitle = {Proceedings of the AAAI Conference on Artificial Intelligence},
  volume = {39},
  number = {12},
  pages = {12818--12826},
  year = {2025},
  journal = {Proceedings of the AAAI Conference on Artificial Intelligence},
  doi = {10.1609/aaai.v39i12.33398},
  publisher = {Association for the Advancement of Artificial Intelligence (AAAI)},
}

@inproceedings{reimers-gurevych-2019-sentence,
  title = {Sentence-{BERT}: Sentence Embeddings using {S}iamese {BERT}-Networks},
  author = {Reimers, Nils  and Gurevych, Iryna},
  editor = {Inui, Kentaro  and Jiang, Jing  and Ng, Vincent  and Wan, Xiaojun},
  booktitle = {Conference on Empirical Methods in Natural Language Processing},
  year = {2019},
  journal = {Conference on Empirical Methods in Natural Language Processing},
  pages = {3980-3990},
  doi = {10.18653/v1/D19-1410},
  publisher = {Association for Computational Linguistics},
}

@inproceedings{kim2024improving,
  title = {Improving multi-hop logical reasoning in knowledge graphs with context-aware query representation learning},
  author = {Kim, Jeonghoon and Jung, Heesoo and Jang, Hyeju and Park, Hogun},
  booktitle = {Findings of the Association for Computational Linguistics: ACL 2024},
  pages = {15978--15991},
  year = {2024},
  journal = {Findings of the Association for Computational Linguistics ACL 2024},
  doi = {10.18653/v1/2024.findings-acl.946},
  publisher = {Association for Computational Linguistics},
}

@inproceedings{tan2025paths,
  title = {Paths-over-graph: Knowledge graph empowered large language model reasoning},
  author = {Tan, Xingyu and Wang, Xiaoyang and Liu, Qing and Xu, Xiwei and Yuan, Xin and Zhang, Wenjie},
  booktitle = {Proceedings of the ACM on Web Conference 2025},
  pages = {3505--3522},
  year = {2025},
  journal = {Proceedings of the ACM on Web Conference 2025},
  doi = {10.1145/3696410.3714892},
  publisher = {ACM},
}

@inproceedings{chen2024llm,
  title = {LLM-based multi-hop question answering with knowledge graph integration in evolving environments},
  author = {Chen, Ruirui and Jiang, Weifeng and Qin, Chengwei and Rawal, Ishaan Singh and Tan, Cheston and Choi, Dongkyu and Xiong, Bo and Ai, Bo},
  booktitle = {Findings of the Association for Computational Linguistics: EMNLP 2024},
  pages = {14438--14451},
  year = {2024},
  journal = {Findings of the Association for Computational Linguistics: EMNLP 2024},
  doi = {10.18653/v1/2024.findings-emnlp.844},
  publisher = {Association for Computational Linguistics},
}

@article{jin2025search,
  title={Search-r1: Training llms to reason and leverage search engines with reinforcement learning},
  author={Jin, Bowen and Zeng, Hansi and Yue, Zhenrui and Yoon, Jinsung and Arik, Sercan and Wang, Dong and Zamani, Hamed and Han, Jiawei},
  journal={arXiv preprint arXiv:2503.09516},
  year={2025}
}

@article{gandhi2025cognitive,
  title = {Cognitive behaviors that enable self-improving reasoners, or, four habits of highly effective stars},
  author = {Gandhi, Kanishk and Chakravarthy, Ayush and Singh, Anikait and Lile, Nathan and Goodman, Noah D},
  journal = {arXiv.org},
  year = {2025},
  doi = {10.48550/arXiv.2503.01307},
}

@inproceedings{yang2018hotpotqa,
  title = {HotpotQA: A dataset for diverse, explainable multi-hop question answering},
  author = {Yang, Zhilin and Qi, Peng and Zhang, Saizheng and Bengio, Yoshua and Cohen, William and Salakhutdinov, Ruslan and Manning, Christopher D},
  booktitle = {Conference on Empirical Methods in Natural Language Processing},
  pages = {2369--2380},
  year = {2018},
  journal = {Conference on Empirical Methods in Natural Language Processing},
  doi = {10.18653/v1/D18-1259},
  publisher = {Association for Computational Linguistics},
}

@misc{trivedi2022musique,
      title={MuSiQue: Multihop Questions via Single-hop Question Composition}, 
      author={Harsh Trivedi and Niranjan Balasubramanian and Tushar Khot and Ashish Sabharwal},
      year={2022},
      eprint={2108.00573},
      archivePrefix={arXiv},
      primaryClass={cs.CL},
      url={https://arxiv.org/abs/2108.00573}, 
}

@inproceedings{chepurova2026wikontic,
  title = {Wikontic: Constructing Wikidata-Aligned, Ontology-Aware Knowledge Graphs with Large Language Models},
  author = {Chepurova, Alla and Bulatov, Aydar and Burtsev, Mikhail and Kuratov, Yurii},
  booktitle = {Proceedings of the 19th Conference of the European Chapter of the Association for Computational Linguistics (Volume 1: Long Papers)},
  pages = {8304--8319},
  year = {2026},
  journal = {Proceedings of the 19th Conference of the European Chapter of the Association for Computational Linguistics (Volume 1: Long Papers)},
  doi = {10.18653/v1/2026.eacl-long.388},
  publisher = {Association for Computational Linguistics},
}

@inproceedings{zhang2024extract,
    title = "Extract, Define, Canonicalize: An {LLM}-based Framework for Knowledge Graph Construction",
    author = "Zhang, Bowen  and
      Soh, Harold",
    editor = "Al-Onaizan, Yaser  and
      Bansal, Mohit  and
      Chen, Yun-Nung",
    booktitle = "Proceedings of the 2024 Conference on Empirical Methods in Natural Language Processing",
    month = nov,
    year = "2024",
    address = "Miami, Florida, USA",
    publisher = "Association for Computational Linguistics",
    url = "https://aclanthology.org/2024.emnlp-main.548/",
    doi = "10.18653/v1/2024.emnlp-main.548",
    pages = "9820--9836"
}

@article{bai2025autoschemakg,
  title = {Autoschemakg: Autonomous knowledge graph construction through dynamic schema induction from web-scale corpora},
  author = {Bai, Jiaxin and Fan, Wei and Hu, Qi and Zong, Qing and Li, Chunyang and Tsang, Hong Ting and Luo, Hongyu and Yim, Yauwai and Huang, Haoyu and Zhou, Xiao and others},
  journal = {Proceedings of the 64th Annual Meeting of the Association for Computational Linguistics (Volume 1: Long Papers)},
  year = {2026},
  pages = {20557-20584},
  doi = {10.18653/v1/2026.acl-long.942},
  publisher = {Association for Computational Linguistics},
}

@inproceedings{izacard2021leveraging,
  title = {Leveraging passage retrieval with generative models for open domain question answering},
  author = {Izacard, Gautier and Grave, Edouard},
  booktitle = {Conference of the European Chapter of the Association for Computational Linguistics},
  pages = {874--880},
  year = {2020},
  journal = {Conference of the European Chapter of the Association for Computational Linguistics},
  doi = {10.18653/v1/2021.eacl-main.74},
  publisher = {Association for Computational Linguistics},
}

@article{schick2023toolformer,
  title = {Toolformer: Language models can teach themselves to use tools},
  author = {Schick, Timo and Dwivedi-Yu, Jane and Dessì, Roberto and Raileanu, R. and Lomeli, M. and Zettlemoyer, Luke and Cancedda, Nicola and Scialom, Thomas},
  journal = {Advances in Neural Information Processing Systems 36},
  volume = {36},
  pages = {68539--68551},
  year = {2023},
  doi = {10.52202/075280-2997},
  publisher = {Neural Information Processing Systems Foundation, Inc. (NeurIPS)},
}

@inproceedings{yao2022react,
  title = {React: Synergizing reasoning and acting in language models},
  author = {Yao, Shunyu and Zhao, Jeffrey and Yu, Dian and Du, Nan and Shafran, Izhak and Narasimhan, Karthik R and Cao, Yuan},
  booktitle = {International Conference on Learning Representations},
  year = {2022},
  journal = {International Conference on Learning Representations},
}

@inproceedings{zheng2025deepresearcher,
  title = {Deepresearcher: Scaling deep research via reinforcement learning in real-world environments},
  author = {Zheng, Yuxiang and Fu, Dayuan and Hu, Xiangkun and Cai, Xiaojie and Ye, Lyumanshan and Lu, Pengrui and Liu, Pengfei},
  booktitle = {Proceedings of the 2025 Conference on Empirical Methods in Natural Language Processing},
  pages = {414--431},
  year = {2025},
  journal = {Proceedings of the 2025 Conference on Empirical Methods in Natural Language Processing},
  doi = {10.18653/v1/2025.emnlp-main.22},
  publisher = {Association for Computational Linguistics},
}

@article{chen2025learning,
  title = {Learning to reason with search for llms via reinforcement learning},
  author = {Chen, Mingyang and Li, Tianpeng and Sun, Haoze and Zhou, Yijie and Zhu, Chenzheng and Wang, Haofen and Pan, Jeff Z. and Zhang, Wen and Chen, Hua-zeng and Yang, Fan and others},
  journal = {Advances in Neural Information Processing Systems 38},
  year = {2025},
  pages = {94974-94994},
  doi = {10.52202/085713-2858},
  publisher = {Neural Information Processing Systems Foundation, Inc. (NeurIPS)},
}

@article{chang2026karl,
  title = {KARL: Knowledge Agents via Reinforcement Learning},
  author = {Chang, Jonathan D. and Drozdov, Andrew and Toshniwal, Shubham and Oertell, Owen and Trott, Alex and Portes, J. and Gupta, Abhay and Koppol, Pallavi and Baheti, Ashutosh and Kulinski, Sean and others},
  journal = {arXiv.org},
  year = {2026},
  doi = {10.48550/arXiv.2603.05218},
}

@inproceedings{devlin2019bert,
  title = {Bert: Pre-training of deep bidirectional transformers for language understanding},
  author = {Devlin, Jacob and Chang, Ming-Wei and Lee, Kenton and Toutanova, Kristina},
  booktitle = {North American Chapter of the Association for Computational Linguistics},
  pages = {4171--4186},
  year = {2019},
  journal = {North American Chapter of the Association for Computational Linguistics},
  doi = {10.18653/v1/N19-1423},
}

@inproceedings{petroni2019language,
  title = {Language models as knowledge bases?},
  author = {Petroni, F. and Rocktäschel, Tim and Lewis, Patrick and Bakhtin, A. and Wu, Yuxiang and Miller, Alexander H. and Riedel, Sebastian},
  booktitle = {Conference on Empirical Methods in Natural Language Processing},
  pages = {2463--2473},
  year = {2019},
  journal = {Conference on Empirical Methods in Natural Language Processing},
  doi = {10.18653/v1/D19-1250},
  publisher = {Association for Computational Linguistics},
}

@article{allen2023physics,
  title = {Physics of language models: Part 3.2, knowledge manipulation},
  author = {Allen-Zhu, Zeyuan and Li, Yuanzhi},
  journal = {International Conference on Learning Representations},
  year = {2023},
  doi = {10.2139/ssrn.5250621},
  publisher = {Elsevier BV},
}

@inproceedings{kandpal2023large,
  title = {Large language models struggle to learn long-tail knowledge},
  author = {Kandpal, Nikhil and Deng, Haikang and Roberts, Adam and Wallace, Eric and Raffel, Colin},
  booktitle = {International conference on machine learning},
  pages = {15696--15707},
  year = {2022},
  organization = {PMLR},
  journal = {International Conference on Machine Learning},
}

@article{yang2025qwen3,
  title = {Qwen3 technical report},
  author = {Yang, An and Li, Anfeng and Yang, Baosong and Zhang, Beichen and Hui, Binyuan and Zheng, Bo and Yu, Bowen and Gao, Chang and Huang, Chengen and Lv, Chenxu and others},
  journal = {arXiv},
  year = {2025},
}

@article{shao2024deepseekmath,
  title = {Deepseekmath: Pushing the limits of mathematical reasoning in open language models},
  author = {Shao, Zhihong and Wang, Peiyi and Zhu, Qihao and Xu, R. and Song, Jun-Mei and Zhang, Mingchuan and Li, Y. K. and Wu, Yu and Guo, Daya},
  journal = {arXiv.org},
  year = {2024},
  doi = {10.48550/arXiv.2402.03300},
}

@article{gao2025beyond,
  title = {Beyond ten turns: Unlocking long-horizon agentic search with large-scale asynchronous rl},
  author = {Gao, Jiaxuan and Fu, Wei and Xie, Minyang and Xu, Shusheng and He, Chuyi and Mei, Zhiyu and Zhu, Banghua and Wu, Yi},
  journal = {arXiv.org},
  year = {2025},
  doi = {10.48550/arXiv.2508.07976},
}

@article{dedhia2025bottom,
  title = {Bottom-up Domain-specific Superintelligence: A Reliable Knowledge Graph is What We Need},
  author = {Dedhia, Bhishma and Kansal, Yuval and Jha, Niraj K},
  journal = {arXiv.org},
  year = {2025},
  doi = {10.48550/arXiv.2507.13966},
}

@article{kansal2026knowledge,
  title = {Knowledge Graphs are Implicit Reward Models: Path-Derived Signals Enable Compositional Reasoning},
  author = {Kansal, Yuval and Jha, Niraj K},
  journal = {arXiv.org},
  year = {2026},
  doi = {10.48550/arXiv.2601.15160},
}

@inproceedings{zhong2023mquake,
  title = {Mquake: Assessing knowledge editing in language models via multi-hop questions},
  author = {Zhong, Zexuan and Wu, Zhengxuan and Manning, Christopher D and Potts, Christopher and Chen, Danqi},
  booktitle = {Proceedings of the 2023 Conference on Empirical Methods in Natural Language Processing},
  pages = {15686--15702},
  year = {2023},
  journal = {Proceedings of the 2023 Conference on Empirical Methods in Natural Language Processing},
  doi = {10.18653/v1/2023.emnlp-main.971},
  publisher = {Association for Computational Linguistics},
}

@article{vrandevcic2014wikidata,
  title={Wikidata: a free collaborative knowledgebase},
  author={Vrande{\v{c}}i{\'c}, Denny and Kr{\"o}tzsch, Markus},
  journal={Communications of the ACM},
  volume={57},
  number={10},
  pages={78--85},
  year={2014},
  publisher={ACM New York, NY, USA}
}

@inproceedings{saxena2020improving,
  title = {Improving multi-hop question answering over knowledge graphs using knowledge base embeddings},
  author = {Saxena, Apoorv and Tripathi, Aditay and Talukdar, Partha},
  booktitle = {Annual Meeting of the Association for Computational Linguistics},
  pages = {4498--4507},
  year = {2020},
  journal = {Annual Meeting of the Association for Computational Linguistics},
  doi = {10.18653/v1/2020.acl-main.412},
  publisher = {Association for Computational Linguistics},
}

@inproceedings{ho2020constructing,
  title = {Constructing a multi-hop qa dataset for comprehensive evaluation of reasoning steps},
  author = {Ho, Xanh and Nguyen, Anh-Khoa Duong and Sugawara, Saku and Aizawa, Akiko},
  booktitle = {International Conference on Computational Linguistics},
  pages = {6609--6625},
  year = {2020},
  journal = {International Conference on Computational Linguistics},
  doi = {10.18653/V1/2020.COLING-MAIN.580},
  publisher = {International Committee on Computational Linguistics},
}

@inproceedings{barnett2024seven,
  title = {Seven failure points when engineering a retrieval augmented generation system},
  author = {Barnett, Scott and Kurniawan, Stefanus and Thudumu, Srikanth and Brannelly, Zach and Abdelrazek, Mohamed},
  booktitle = {2024 IEEE/ACM 3rd International Conference on AI Engineering – Software Engineering for AI (CAIN)},
  pages = {194--199},
  year = {2024},
  journal = {2024 IEEE/ACM 3rd International Conference on AI Engineering – Software Engineering for AI (CAIN)},
  doi = {10.1145/3644815.3644945},
  publisher = {ACM},
}

@inproceedings{shi2023large,
  title = {Large language models can be easily distracted by irrelevant context},
  author = {Shi, Freda and Chen, Xinyun and Misra, Kanishka and Scales, Nathan and Dohan, David and Chi, Ed H. and Scharli, Nathanael and Zhou, Denny},
  booktitle = {International Conference on Machine Learning},
  pages = {31210--31227},
  year = {2023},
  organization = {PMLR},
  journal = {International Conference on Machine Learning},
  doi = {10.48550/arXiv.2302.00093},
}

@inproceedings{karpukhin2020dense,
  title = {Dense passage retrieval for open-domain question answering},
  author = {Karpukhin, Vladimir and Oguz, Barlas and Min, Sewon and Lewis, Patrick and Wu, Ledell and Edunov, Sergey and Chen, Danqi and Yih, Wen-tau},
  booktitle = {Conference on Empirical Methods in Natural Language Processing},
  pages = {6769--6781},
  year = {2020},
  journal = {Conference on Empirical Methods in Natural Language Processing},
  doi = {10.18653/v1/2020.emnlp-main.550},
  publisher = {Association for Computational Linguistics},
}

@inproceedings{jin2025flashrag,
  title = {Flashrag: A modular toolkit for efficient retrieval-augmented generation research},
  author = {Jin, Jiajie and Zhu, Yutao and Yang, Xinyu and Zhang, Chenghao and Dou, Zhicheng},
  booktitle = {The Web Conference},
  pages = {737--740},
  year = {2024},
  journal = {The Web Conference},
  doi = {10.1145/3701716.3715313},
  publisher = {ACM},
}

@inproceedings{li2025search,
  title = {Search-o1: Agentic search-enhanced large reasoning models},
  author = {Li, Xiaoxi and Dong, Guanting and Jin, Jiajie and Zhang, Yuyao and Zhou, Yujia and Zhu, Yutao and Zhang, Peitian and Dou, Zhicheng},
  booktitle = {Proceedings of the 2025 Conference on Empirical Methods in Natural Language Processing},
  pages = {5420--5438},
  year = {2025},
  journal = {Proceedings of the 2025 Conference on Empirical Methods in Natural Language Processing},
  doi = {10.18653/v1/2025.emnlp-main.276},
  publisher = {Association for Computational Linguistics},
}

@article{lewis2020retrieval,
  title = {Retrieval-augmented generation for knowledge-intensive nlp tasks},
  author = {Lewis, Patrick and Perez, Ethan and Piktus, Aleksandara and Petroni, F. and Karpukhin, Vladimir and Goyal, Naman and Kuttler, Heinrich and Lewis, M. and Yih, Wen-tau and Rocktäschel, Tim and others},
  journal = {Neural Information Processing Systems},
  volume = {33},
  pages = {9459--9474},
  year = {2020},
}

@inproceedings{trivedi2023interleaving,
  title = {Interleaving retrieval with chain-of-thought reasoning for knowledge-intensive multi-step questions},
  author = {Trivedi, Harsh and Balasubramanian, Niranjan and Khot, Tushar and Sabharwal, Ashish},
  booktitle = {Proceedings of the 61st annual meeting of the association for computational linguistics (volume 1: long papers)},
  pages = {10014--10037},
  year = {2023},
  journal = {Proceedings of the 61st Annual Meeting of the Association for Computational Linguistics (Volume 1: Long Papers)},
  doi = {10.18653/v1/2023.acl-long.557},
  publisher = {Association for Computational Linguistics},
}

@article{yue2025does,
  title = {Does reinforcement learning really incentivize reasoning capacity in llms beyond the base model?},
  author = {Yue, Yang and Chen, Zhiqi and Lu, Rui and Zhao, Andrew and Wang, Zhaokai and Song, Shiji and Huang, Gao},
  journal = {Advances in Neural Information Processing Systems 38},
  year = {2025},
  pages = {64304-64339},
  doi = {10.52202/085713-1933},
  publisher = {Neural Information Processing Systems Foundation, Inc. (NeurIPS)},
}

@article{comanici2025gemini,
  title = {Gemini 2.5: Pushing the frontier with advanced reasoning, multimodality, long context, and next generation agentic capabilities},
  author = {Comanici, Gheorghe and Bieber, E. and Schaekermann, Mike and Pasupat, Ice and Sachdeva, Noveen and Dhillon, Inderjit S. and Blistein, Marcel and Ram, Ori and Zhang, Dan and Rosen, Evan and others},
  journal = {arXiv.org},
  year = {2025},
}

@inproceedings{bi2026parameters,
  title={Parameters vs. context: Fine-grained control of knowledge reliance in language models},
  author={Bi, Baolong and Liu, Shenghua and Wang, Yiwei and Xu, Yilong and Fang, Junfeng and Mei, Lingrui and Cheng, Xueqi},
  booktitle={International Conference on Learning Representations},
  volume={2026},
  pages={106786--106808},
  year={2026}
}
